%% file: main.tex
\documentclass[10pt,twocolumn,letterpaper]{article}

\usepackage[pagenumbers]{iccv} 
\usepackage{times}
\usepackage{epsfig}
\usepackage{graphicx}
\usepackage{amsmath, amsfonts, amssymb}
\usepackage{algorithm, algpseudocode} %
\usepackage{enumitem}
\usepackage{wrapfig}
\usepackage{pifont} %
\usepackage{stmaryrd} %
\usepackage{booktabs, multirow} %
\usepackage{tablefootnote}
\usepackage{comment}
\usepackage{colortbl} %
\usepackage{xcolor}
\usepackage{sidecap} %
\usepackage{dblfloatfix} %
\usepackage{colortbl}
\usepackage{bm}
\usepackage{arydshln}
\usepackage{float}
\usepackage{pgfplots}
\usepackage[utf8]{inputenc}
\usepackage{xcolor}
\usepackage{caption}
\usepackage{array}       
\usepackage{graphicx}     
\usepackage{multirow}     
\usepackage{booktabs}     
\usepackage{adjustbox}    
\usepackage{amssymb} 
\usepackage{placeins} 
\FloatBarrier

\usepackage{capt-of} 
\pgfplotsset{compat=1.17}
\definecolor{lightblue}{RGB}{202, 240, 248} 
\definecolor{lightcream}{RGB}{245, 235, 224}
\definecolor{lightgreen}{RGB}{253, 252, 220}

\usepackage{amsmath} 


\definecolor{iccvblue}{rgb}{0.21,0.49,0.74}
\usepackage[pagebackref,breaklinks,colorlinks,allcolors=iccvblue]{hyperref}

\def\paperID{2068} 
\def\confName{ICCV}
\def\confYear{2025}

\title{Your Text Encoder Can Be An Object-Level Watermarking Controller}

\author{
    \textbf{Naresh Kumar Devulapally}$^{1}$ \quad 
    \textbf{Mingzhen Huang}$^{1}$ \quad
    \textbf{Vishal Asnani}$^{2}$ \\
    \textbf{Shruti Agarwal}$^{2}$ \quad 
    \textbf{Siwei Lyu}$^{1}$ \quad 
    \textbf{Vishnu Suresh Lokhande}$^{1}$ \\
    \vspace{0.1em}
    $^{1}$University at Buffalo, SUNY \quad 
    $^{2}$Adobe Research \\
    {\tt\small \{devulapa, vishnulo, mhuang33, siweilyu\}@buffalo.edu} \quad
    {\tt\small \{shragarw, vasnani\}@adobe.com}
}

\begin{document}


\maketitle

\input{sec/0_abstract}
\input{sec/1_intro}

\input{sec/2_related_work}

\input{sec/3_method}

\input{sec/5_experiments} 
\input{sec/6_conclusion}

{
    \small
    \bibliographystyle{ieeenat_fullname}
    \bibliography{main}
}

\input{sec/X_suppl}

%


\end{document}

%% file: sec/0_abstract.tex

\begin{abstract}

\vspace{-3mm}

Invisible watermarking of AI-generated images can help with copyright protection, enabling detection and identification of AI-generated media. In this work, we present a novel approach to watermark images of T2I Latent Diffusion Models (LDMs). By only fine-tuning text token embeddings $\mathcal{W}_*$, we enable watermarking in selected objects or parts of the image, offering greater flexibility compared to traditional full-image watermarking. Our method leverages the text encoder’s compatibility across various LDMs, allowing plug-and-play integration for different LDMs. Moreover, introducing the watermark early in the encoding stage improves robustness to adversarial perturbations in later stages of the pipeline. Our approach achieves $99\%$ bit accuracy ($48$ bits) with a $10^5 \times$ reduction in model parameters, enabling efficient watermarking. Code: \href{https://github.com/naresh-ub/localize-watermark}{GitHub}.

\end{abstract}

%% file: sec/1_intro.tex
\section{Introduction} \label{sec:intro}

As debates around copyright and intellectual property intensify, the need for effective watermarking solutions increases \citep{wen2024tree}. With generative AI models producing indistinguishable images from human-made art, maintaining authorship provenance in digital media is crucial. Efforts are underway, from legislative measures like the Executive Order on Safe, Secure, and Trustworthy Artificial Intelligence \citep{biden2023executive}, to industry initiatives that aim to watermark all AI-generated content \citep{noti2024regulating} \citep{smuha2021eu}. These developments highlight the growing importance of watermarking as a key area of study.

As a result, multiple Generative AI (GenAI) watermarking methods have been proposed~\cite{haden2023now,  wen2024tree, 10377226} to invisibly watermark the entire image while generating the image. However, lots of current watermark methods~\cite{pmlr-v235-feng24k, bui2023rostealsrobuststeganographyusing} can only encode a watermark to an open-box generative model as they need to access the latent space.
Moreover, similar to other invisible watermarking techniques~\cite{bui2023trustmarkuniversalwatermarkingarbitrary, 10377226, pmlr-v235-feng24k}, these methods also suffer with the inherent trade-off between quality and robustness of the embedded signal. Owing to the high imperceptibility requirements for the high-quality image generation, these watermarks more often lacks robustness~\citep{haden2023now, zhao2024invisibleimagewatermarksprovably}. To improve the watermark imperceptibility, partial watermarking has been proposed to limit the watermark related changes only to selected ``non-salient" objects or regions in the image~\cite{nikolaidis2001region}. In our paper, we bring such partial watermarking to text-to-image generation pipeline. Our partial watermarking is not only for improving the quality of the watermarked image, but also to give the user the flexibility to watermark only selected object of interest in the generated image without accessing the latent space. This is crucial for scenarios where the user would like to protect the unique object in the image, see ~\cref{fig:teaser}.


Specifically, we propose a token-based watermarking approach that can embed watermarks into selected objects or partial regions of an image. Unlike previous works that have explored architectural modifications, such as the addition of secret encoders \citep{bui2023rostealsrobuststeganographyusing}, distortion layers \citep{pmlr-v235-feng24k}, and adapters \citep{feng2023catcheverywhereguardingtextual}, our watermarking method focuses solely on learning a watermark embedding. This allows for a \textit{prêt-à-porter} style of training \citep{rout2024rbmodulationtrainingfreepersonalizationdiffusion}, leveraging pre-trained models with minimal adjustments. Additionally, compared with prior works~\cite{bui2023rostealsrobuststeganographyusing, feng2023catcheverywhereguardingtextual, pmlr-v235-feng24k}, our model is blind, meaning that the entire generative model is neither modified nor requires additional training. A simple watermark token is learned by our model and can be plugged into any diffusion model for watermark generation. Previous works~\citep{10377226, wen2024tree} have investigated image watermarking without altering the architecture. Although effective, they lack the convenience of \textit{prêt-à-porter} training, which offers the key advantage of personalizing large models with minimal retraining while using the model's core components as-is. In contrast, our approach enables users to apply watermarking directly at the prompt stage without modifying core model components such as the UNet or image decoder. We introduce new pseudo-tokens $\bm{\mathcal{W}_*}$ into the model's vocabulary, where the $\bm{\mathcal{W}_*}$ can be used as an input text prompt to be applied on any T2I diffusion models for generation watermarked images.
This also enables users to selectively watermark specific image regions by leveraging the cross-attention between the special token and visual region in the image where the watermark needs to be embedded. By integrating watermarking functionality early in the text-to-image pipeline, our approach performs in-generation watermarking, offering greater robustness against image manipulation attacks compared to post-processing techniques while improving the overall quality of the watermarked images. 

We make the following \textbf{key contributions}:
\begin{enumerate}
    \item We introduce a novel pseudo-token with watermarking capabilities, learning a new embedding vector for it. We investigate the impact of applying this pseudo-token conditioning at different timesteps of the diffusion process, finding that timesteps closer to the VAE encoder in LDMs during forward process offer enhanced image generation quality.
    \item We introduce object-level watermarking, allowing for selective watermarking with greater precision than traditional whole-image approaches.
    \item We offer plug-and-play integration with various Stable Diffusion (SD) variants by embedding the watermark directly via the text encoder. Our approach achieves $99\%$ bit accuracy with a $10^5 \times$ reduction in model parameters, enabling efficient watermarking with a throughput of $48$ bits.
\end{enumerate}

%% file: sec/2_related_work.tex
\section{Related Works}
\label{sec:related_works}

{\bf Text-Guided Image Generation Diffusion Models:} Diffusion models have revolutionized text-to-image (T2I) generation, surpassing GANs in fidelity and diversity. By iteratively denoising random noise into images conditioned on text prompts, these models enable fine-grained control over generation. Frameworks like DALL-E 2, Imagen~\citep{imagenteamgoogle2024imagen3}, and Stable Diffusion~\citep{song2022denoisingdiffusionimplicitmodels} democratize high-quality image synthesis, while diffusion-based editing methods~\citep{huang2024paralleleditsefficientmultiaspecttextdriven} enable localized modifications. However, their widespread adoption raises challenges, including copyright infringement and harmful content generation, necessitating controllable synthesis mechanisms such as watermarking.

\noindent {\bf Textual Inversion for Personalized Image Generation:} Textual Inversion~\citep{gal2022imageworthwordpersonalizing} embeds visual or stylistic concepts into T2I models through learnable tokens, enabling personalization without altering model parameters. This lightweight approach has been extended to style transfer~\citep{zhang2023inversion}. In this work, we find that training the special token not only allows for flexible concept learning but also could acts as a watermark controller.

\noindent {\bf Constraint-Based Control in Text-to-Image Generation:} Text-to-image generation has gained significant popularity, especially to incorporate constraints that enhance control over generated content. Common constraints include spatial or layout constraints, which dictate object placement and dimensions within the generated image to meet predefined spatial requirements \citep{liang2024diffusion4dfastspatialtemporalconsistent}. Another type, multi-view consistency constraints, ensures scene consistency across perspectives, preserving layout, lighting, and depth even in complex views like outdoor or stylized scenes \citep{zhou2023sceneconditional3dobjectstylization}. Attribute preservation constraints maintain prompt-specified attributes (e.g., color, texture, size) in the generated image, ensuring semantic alignment. Finally, watermarking constraints embed an invisible watermark to verify ownership while preserving image quality. Watermarks can be embedded through polytope constraints defined by orthogonal Gaussian vectors or high-frequency components, allowing watermarking without image degradation. Constraint-based generation methods, which offer explicit control over outputs, draw inspiration from classifier training techniques that optimize using soft penalties \citep{daras2022softdiffusionscorematching}. Recent approaches such as \citep{ho2022classifierfreediffusionguidance} have achieved notable success with convergence guarantees and adaptability across data regimes .

\noindent {\bf  In-generation Image watermarking in T2I generation:} With the increasing popularity of diffusion models, recent research has explored embedding watermarks within the diffusion process, either by manipulating the noise schedule to encode watermark information or by conditioning the model to output watermarked images. Tree-ring~\cite{wen2024tree} proposed to encode a watermark into the noise space, it can be simply detected by applying a DDIM \cite{song2022denoisingdiffusionimplicitmodels} inversion to obtain the noise.
However, current in-generation watermark methods \cite{rezaei2024lawausinglatentspace, 10377226, pmlr-v235-feng24k} cannot inject a watermark in a local region, \eg an object. As many diffusion-based image editing method have been emerging for a long time, the object-level watermarking is long overdue. Recently, WAM ~\cite{sander2024watermarklocalizedmessages} discusses localized watermarking and proposes a training mechanism for watermarked region detection from watermarked image. However, WAM still requires segmentation masks to \textit{localize} watermark during watermark embedding in both training and inference. 

\noindent {\bf Blind v/s Non-Blind Watermarking:} Mentioned in \cite{li2021surveydeepneuralnetwork}, blind detection methods verify ownership using only the watermarked image, eliminating reliance on auxiliary metadata. Our work adopts this practical approach, ensuring compatibility with standard detection workflows.

\noindent \noindent Motivated by these insights, we explore similar methods in text-to-image generation, where constraints are often enforced as soft penalties via gradient-based optimization. Current diffusion-based watermarking methods lack spatial control, limiting object-centric applications. Recent advances in localized diffusion editing~\citep{huang2024paralleleditsefficientmultiaspecttextdriven} remain unexplored for watermark integration. By unifying textual inversion with constraint-based control, we propose the first framework for object-level in-generation watermarking in T2I generation, addressing traceability and granularity.

%% file: sec/3_method.tex
\begin{figure*}[!t]
    \centering
    \includegraphics[width=\textwidth]{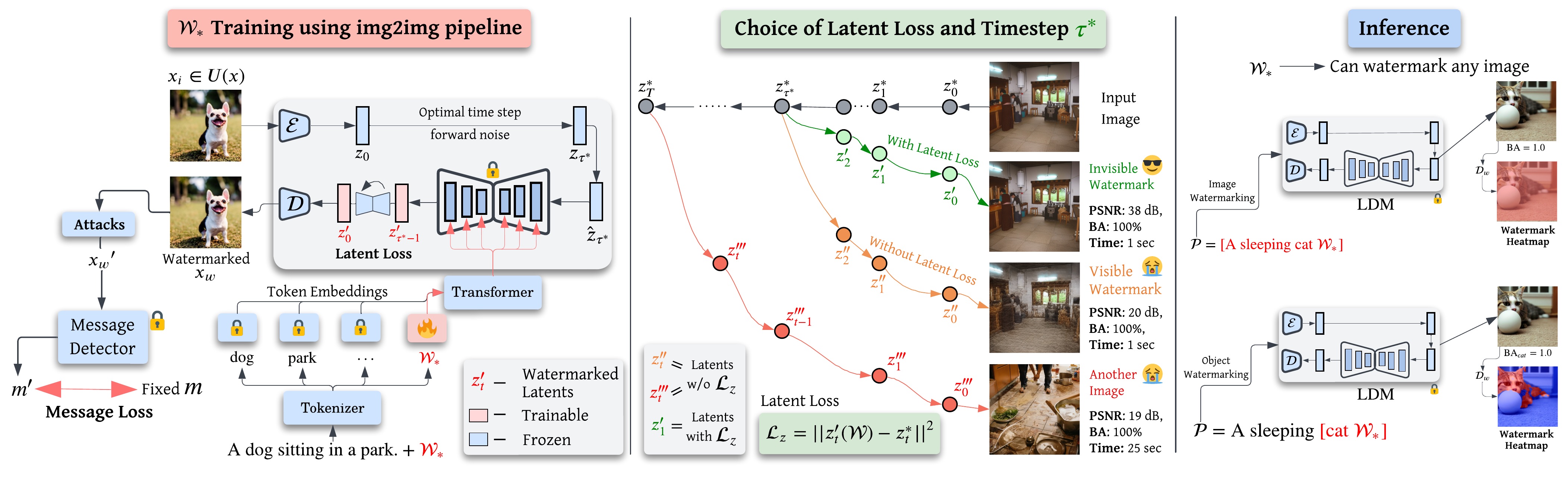}
    \caption{\textbf{$\bm{\mathcal{W}_*}$ training pipeline.} \emph{(Left) To find $\mathcal{W}_*$ token embeddings, we use an Img2Img generation pipeline. $\mathcal{D}$ and $\mathcal{D}_w$ represent VAE decoder in the LDM, and Watermark Detector respectively. While training, we send the input image through LDM encoder to retrieve the latent $z_0$, we then add a forward diffusion noise of $\tau^*$ timesteps, followed by iteratively denoising $z_{\tau^*}$ using Classifier-Free Guidance \citep{ho2022classifierfreediffusionguidance} from $[\tau^* \to 0]$ to retrieve $z'_{0,w}$. During the denoising process, we train for $\mathcal{W}_*$ token embeddings. (Middle) We use latent matching loss to control the trajectory of watermarked latents and bit loss to find $\mathcal{W}_*$ token embeddings. (Right) We then use trained $\mathcal{W}_*$ embeddings to generate watermarked images.}}
    \label{fig:method}
\end{figure*}

\section{Problem Statement}

Our problem setting uses Latent Diffusion Models (LDMs)~\cite{Rombach_2022_CVPR} for image generation. We consider an in-generation watermarking scenario, as defined in \cref{sec:related_works}, where the aim is to utilize existing modules within the LDM pipeline to watermark images while generation.

\noindent \textbf{Watermark Embedder:} Given an input text prompt ($\mathcal{P}=\{p_0, p_1, ... p_n\}$) and/or an image $\mathcal{I}$, the aim of an in-generation watermark embedder ($W_e$) is to generate a watermarked image $\mathcal{I}_w = LDM(\mathcal{I} \mid W_e, \mathcal{P}$). Each embedder $W_e$ is associated with a watermark key $(m)$ (a bit string containing $0$, $1$ similar with prior works~\cite{10377226,pmlr-v235-feng24k}) of length $k$, i.e., $m \in \{0,1\}^k$. From here we shall use $W_{e,m}$ to denote a watermark embedder and $\mathcal{I}_{w,m}$ to denote a watermarked image.


\noindent \textbf{Watermark Detector:} Given a watermarked image $\mathcal{I}_{w,m}$, watermark detector ${D}_w(\cdot)$ is a neural network that predicts the key $m$ from $\mathcal{I}_{w,m}$. Our method utilizes \textit{Blind Watermarking} scenario mentioned in \cref{sec:related_works} where ${D}_w(\cdot)$ only requires the watermarked image $\mathcal{I}_{w,m}$ as input without the need for any additional metadata.

\noindent \textbf{Attack Module:} Attack module $\mathcal{A}(\cdot)$ represents unintentional (or) intentional transforms applied to the watermarked image $\mathcal{I}_{w,m}$ that could result in the loss of watermark key during watermark detection. A successful attack module results in imperfect watermark detection, i.e., ${D}_w(\mathcal{A}(\mathcal{I}_{w,m})) \neq m$. This module can usually be seen in the form of image compression by public platforms or an intentional malicious attacker that aims to remove watermark from $\mathcal{I}_{w,m}$. Resistance to an attack module is a fundamental requirement of a robust watermark embedder $W_e$.

\noindent \textbf{Full-Image and Object-Level Watermarking:} 
Given a text prompt for generation, \textit{full-image watermarking} refers watermarking \textit{entire image} and \textit{object-level watermarking} refers to the scenario that watermarks \textit{specific objects} ${O_i, ... O_j}$ in the image selected by a user. Information about $O_i$ can be provided in various ways. For example \citep{10377226} uses post-processing techniques to localize watermark on various regions using masks. However, using such post-processing methods introduces additional computational overhead and may be susceptible to removal or modification. Instead, integrating object-level watermarking into the generative process enables seamless and robust embedding within the chosen object ${O_i, ... O_j}$ regions.

\noindent Our problem setting considers both image and object-level watermarking scenarios with a constraint that the information for watermarking a specific object $O_i$ shall not be provided in the form of any additional information such as segmentation masks. This information can only be obtained right from the text prompt. Specifically, for our setting, a text prompt $``\text{[A photo of a cat $\mathcal{W}_*$]}"$ denotes full-image watermarking and a text prompt $``\text{A photo of a [cat $\mathcal{W}_*$]}"$ denotes that the object $\text{cat}$ is to be watermarked.

\section{Method} \label{sec:method}

Given an image, $\mathcal{I}$, our training pipeline aims to generate $\mathcal{I}_{w,m}$ where $m$ is the watermark key, $m \in \{0,1\}^k$. We utilize a differentiable watermark detector from \citep{pmlr-v235-feng24k} ${D_w}(I_{w,m})$ that permits gradient flow.

It is known that training larger modules of an LDM pipeline, such as VAE Encoder/Decoder or the UNet, is computationally expensive and do not provide a lightweight, seamless way to integrate watermarking into various other LDM pipelines. There is a need for a unified watermark embedder that can be integrated with various LDM variants.

\subsection{Token Embeddings for Watermarking} Our method utilizes the relatively under-explored Text Encoder of the LDM pipeline \citep{ma2020perceptionorientedsingleimagesuperresolution}. We introduce a new token in the Text Encoder, denoted by $\bm{\mathcal{W}_*}$, and fine-tune the text embeddings of $\bm{\mathcal{W}_*}$ to watermark $\mathcal{I}$. In addition to significant lower parameter requirement compared with prior works \citep{10377226, pmlr-v235-feng24k}, this token $\bm{\mathcal{W}_*}$ act as the watermark trigger that can be seamlessly integrated into the text encoder of \textit{any} LDM pipeline.

Our method initially aims to train $\bm{\mathcal{W}_*}$ token embeddings, similar with Textual Inversion \citep{gal2022imageworthwordpersonalizing}, to generate watermarked latents $z_{t,w} = \epsilon_\theta(z_t,t, \{\mathcal{P}, \mathcal{W}_*\})$. However, we identify the need to \textit{carefully select the optimal noise timesteps} during $\bm{\mathcal{W}*}$ training as different choices impact image quality, watermark robustness and image generation time.

\subsection{Optimal Timestep and Latent Loss} \label{sec:opt_timestep_intro}

We perform empirical studies to observe the effect of different noise timesteps during forward diffusion process while $\bm{\mathcal{W}_*}$ training. As depicted in \cref{fig:method} (middle) (and in the section H. of the supplement), we observe a trade-off between image quality and watermarking performance when choosing different noise timesteps. A relatively large timestep $(t \sim T)$ would degrade the image quality while a relatively small timestep $(t \sim 0) $ would lead a lower bit accuracy but enhanced image generation quality. This observation is consistent with prior findings, as mentioned in \cite{rout2024rbmodulationtrainingfreepersonalizationdiffusion, li2025exploringiterativemanifoldconstraint}, which highlight the impact of noise strength on conditioning fidelity. We identify an optimal timestep $\tau^*=8$ that balances this trade-off between image quality and watermark bit accuracy. During $\bm{\mathcal{W}_*}$ training, we add a forward noise of  $\tau^*$ to the image followed by iterative de-noising to generate watermarked latent $z'_{0, w}$.  $z'_{0, w}$ is then passed into the VAE decoder $Dec(\cdot)$ to generate a watermarked image $I_{w,m}$. $D_w$ takes $I_{w,m}$ as input, ${D_w}(I_{w,m}) = m'$, to train $\bm{\mathcal{W}_*}$ token embeddings on watermark loss $\mathcal{L}_w$. We utilize $BCE$ loss to embed a specific bit key $m$ using $\bm{\mathcal{W}_*}$.

\begin{equation}
    \mathcal{L}_w = BCE(D_w(Dec(z'_{0, w})), m).
\end{equation}

While the optimal timestep $\tau^*$ preserves overall information in $\mathcal{I}$ and significantly reduces the time taken for $I_{w,m}$ generation, we still observe visible corruption in $I_{w,m}$ that hurts imperceptibility watermarking constraint. As a remedy, we employ latent matching loss $\mathcal{L}_z$ that aids our method to perform invisible watermarking.

\begin{equation}
    \mathcal{L}_z = \min_{\mathcal{W}_*} \mathbb{E}_t  \left [ \left\| z^*_{t} - z'_{t}(\mathcal{W}_*) \right\|_2^2 \right ].
\end{equation}

Our method uses $\mathcal{L} = \alpha\mathcal{L}_w + \beta \mathcal{L}_z $ to train $\bm{\mathcal{W}_*}$ token embeddings for watermarking where $\alpha$ and $\beta$ are tunable hyperparameters.

\subsection{Object-level watermarking} \label{sec:obj_wat_method}

\begin{algorithm}[!t]
    \begin{minipage}{\linewidth}
        \caption{Object-level Watermarking during Text-to-Image Generation in Latent Diffusion Models}\label{alg:UAC}
        \begin{algorithmic}[1]
            \Statex \textbf{Input:}\hskip\algorithmicindent LDM Decoder \( \mathcal{D} \); Watermark Token \( \mathcal{W}^i_* \) for object \( i \); Text Prompt \( \mathcal{P} \); overlay strength \( \pi(t) \).
            \State Select objects to watermark, defining the subset \( \{\mathcal{P}^*_0, \mathcal{P}^*_1, \dots, \mathcal{P}^*_k\} \) with corresponding watermarking tokens \( \{\mathcal{W}^*_0, \mathcal{W}^*_1, \dots, \mathcal{W}^*_k\} \) from the text prompt \( \mathcal{P} = \{\mathcal{P}_0, \mathcal{P}_1, \dots, \mathcal{P}_n\} \).
            \State $z_T \in \mathcal{N}(0, I)$
            \For{$t = T, \dots, 0$}               
                \For{each token $\mathcal{P}^*_i \in \{ \mathcal{P}^*_0, \mathcal{P}^*_1, \dots, \mathcal{P}^*_k \}$}
                    \State $\mathcal{M}_{\mathcal{P}^*_i}^{(t)} /  \mathcal{M}_{\mathcal{W}^*_i}^{(t)} \leftarrow$  Attention map of $\mathcal{P}^*_i / \mathcal{W}^*_i$ 
                    \State  $\mathcal{M}_{\mathcal{W}_*}^{(t)}\gets \pi(t) \cdot  \mathcal{M}_{\mathcal{W}_*}^{(t)}$ (Adjust watermark-strength per timestep)
                    \State $\mathcal{M}_{\mathcal{P}_i}^{(t)} \gets (1 - \alpha) \cdot \mathcal{M}_{\mathcal{P}_i}^{(t)} + \alpha \cdot \mathcal{M}_{\mathcal{W}_*}^{(t)}$ (Overlay attention map of $\mathcal{W}^i_*$ on that of $\mathcal{P}^i_*$)
                \EndFor
                
                \State $ \hat\epsilon = \epsilon_\theta(z_t, t, \psi(\mathcal{P}), \mathcal{M}_{\mathcal{P}^*_{0:k}}^{(t)}, \mathcal{M}_{\mathcal{W}^*_{0:k}}^{(t)})$ 
                \State $ z_{t-1} = \sqrt{\frac{\alpha_{t-1}}{\alpha_t}} z_t + \left( \sqrt{\frac{1}{\alpha_{t-1}} - 1} - \sqrt{\frac{1}{\alpha_t} - 1} \right)\hat\epsilon$ 
            \EndFor
            \State \textbf{Output:} Decoder output of$z_0$, that is $\mathcal{D}(z_0)$
            
        \end{algorithmic} 
    \end{minipage}
\end{algorithm}


In the original LDM~\cite{Rombach_2022_CVPR}, the query feature $Q$ and key feature $K$ from cross attention operation defines a token-wise attention mask $M$ where $M = \text{softmax} (Q \cdot K^T) / \sqrt{d})$. As the architecture of the LDM remains unchanged in our model, we inherit this capability to generate corresponding attention mask from text tokens. Such feature is the key in our model unlocking new possibilities for object-level watermarking utilizing these cross-attention maps.

Once the watermarking token embeddings are fine-tuned, we proceed to generate images using a standard text-to-image LDM model, but with an ability to perform object-level watermarking. We utilize cross-attention maps at each timestep $t$ to localize watermark on any chosen object $O_i$ right from the text prompt $\mathcal{P}$ as a token $\mathcal{P}_i ( \in \mathcal{P})$. Specifically, we utilize \citep{hertz2022prompttopromptimageeditingcross} to obtain cross-attention map $M_t$ of $O_i$ given by:


\begin{equation}
    z'_{t-1}, M_{t, O_i} \gets \text{LDM}(z'_t, \mathcal{P}_i, t)
\end{equation}

Given a text prompt with multiple tokens \(\mathcal{P} = \{\mathcal{P}_0, \mathcal{P}_1, \dots, \mathcal{P}_n\}\), the user has the flexibility to choose specific object tokens $\{ \mathcal{P}_i,\mathcal{P}_j, \dots \}$ to watermark. During the image generation process, we extract the cross-attention maps for the objects intended for watermarking from the UNet. These are then combined with the attention maps of the corresponding watermarking tokens $\mathcal{W}_*^i, \mathcal{W}_*^j, \dots $ to precisely localize the watermark on each object allowing us to seamlessly perform multi-object watermarking. 

To bring in the effect of optimal timestep $\tau^*$ (seen during training \cref{sec:opt_timestep_intro}), we introduce a watermark overlay strength controller $\pi(t)$. $\pi(t)$ could be set to a step function with values $0 \enspace (\text{ if } t > \tau^*)$ and $1 \enspace (\text{ if } t \leq \tau^*)$ which exactly mimics the generation scenario seen during training. In addition to the step function, we also test with a smoothing function that brings the effect of $\tau^*$ by giving more weight to timesteps closer to the VAE decoder. At these timesteps we observe increased reliability of the attention maps enhancing the precision of the watermark placement. The complete procedure is outlined in Algorithm \ref{alg:UAC}.

%% file: sec/5_experiments.tex
\vspace{-3mm}

\section{Experiments} %
\label{sec:exp}

\input{sec/tables/table1_watermark_main}




\noindent {\bf Datasets:} We evaluate our method on two datasets, namely, MS-COCO \cite{LinMBHPRDZ14} and WikiArt \cite{artgan2018}. We use a subset of 2,000 images from MS-COCO dataset, similar to \cite{10377226}, for training $\mathcal{W}_*$ token embeddings. Our method is evaluated on 1000 validation captions from MS-COCO and WikiArt each via image-to-image generation and on 100 prompts from \cite{pmlr-v235-feng24k} for text-to-image generation.

\noindent {\bf Evaluation Metrics:} In line with prior work \cite{10377226, pmlr-v235-feng24k}, we assess our watermarking method using the following metrics: 

\textbf{(a)} \textit{Robustness}, measured by Bit Accuracy under various attacks (mentioned in \cref{tab:modified-table-main}). Bit accuracy represents the percentage of correctly detected bits in the watermark key $m$ output from the Detector $D_w(\cdot)$. In addition to Bit Accuracy, we also compare our method with techniques \citep{wen2024tree} that use True Positive Rate as the metrics in supplement (section D). These methods do not embed a bit string $m$.

\textbf{(b)} \textit{Imperceptibility}, measured by Peak Signal-to-Noise Ratio (PSNR), Structural Similarity Index (SSIM) \citep{ma2020perceptionorientedsingleimagesuperresolution, nilsson2020understandingssim}, and Fréchet inception distance (FID)~\cite{jayasumana2024rethinkingfidbetterevaluation}, comparing watermarked and non-watermarked images. 

\textbf{(c)} \textit{Parameter Efficiency}, where our method trains only the textual embeddings, with parameter usage significantly reduced ($10^5 \times$) compared to baselines \cite{10377226, pmlr-v235-feng24k}, as the T2I pipeline and watermark decoder remain frozen.

\noindent {\bf Watermark Detector:} For our experiments we utilize two watermark detectors from \citep{pmlr-v235-feng24k} and \cite{10377226}.



\subsection{Full-Image Watermarking results}

We evaluate full-image watermarking by integrating $\bm{\mathcal{W}_*}$ into the Stable-Diffusion v$1.5$ \cite{song2022denoisingdiffusionimplicitmodels}, using the AquaLora watermark detector \cite{pmlr-v235-feng24k}. Our method's performance is compared to existing watermarking baselines in LDMs, focusing on robustness and imperceptibility. \cref{tab:modified-table-main} presents the results, showing that our approach requires significantly fewer parameters while maintaining high bit accuracy under attacks. We categorize attacks encountered after watermarking into two categories, namely, basic image processing attacks such as Rotation, Resize, Crop, JPEG compression etc as listed in \cref{tab:modified-table-main}, and adversarial attacks such as DiffPure \cite{nie2022diffusionmodelsadversarialpurification}, WMAttacker \cite{zhao2024invisibleimagewatermarksprovably} etc. We provide implementation details of attack module in supplement (section I). 

We consider that robustness to adversarial attacks including Diff-Pure \cite{nie2022diffusionmodelsadversarialpurification} and SDEdit \cite{meng2022sdeditguidedimagesynthesis}, while beneficial, is not essential for watermarking techniques. These attack methods are complex and go beyond common attacks like crop, resize, and rotations. However, our watermarking method integrated into the denoising process demonstrates enhanced robustness to both basic and adversarial attacks outperforming several in-generation watermarking techniques thereby pushing the benchmark for robust watermarking. When compared to AquaLORA \cite{pmlr-v235-feng24k}, while using the same watermark detector, we see an increase in bit accuracy on both common attacks by over $ 20\%$, and over $7 \text{dB}$ in PSNR. Additionally, our method retains high robustness and personalization when used alongside a personalized textual inversion token and/or a personalized UNet.

{\bf Summary.} Our Full-image watermarking achieves improved robustness, lower parameter requirements, and enhanced imperceptibility compared to baselines.

\begin{figure*}[!t] 
\vspace{-0.2in}
    \centering
    \includegraphics[width=\textwidth]{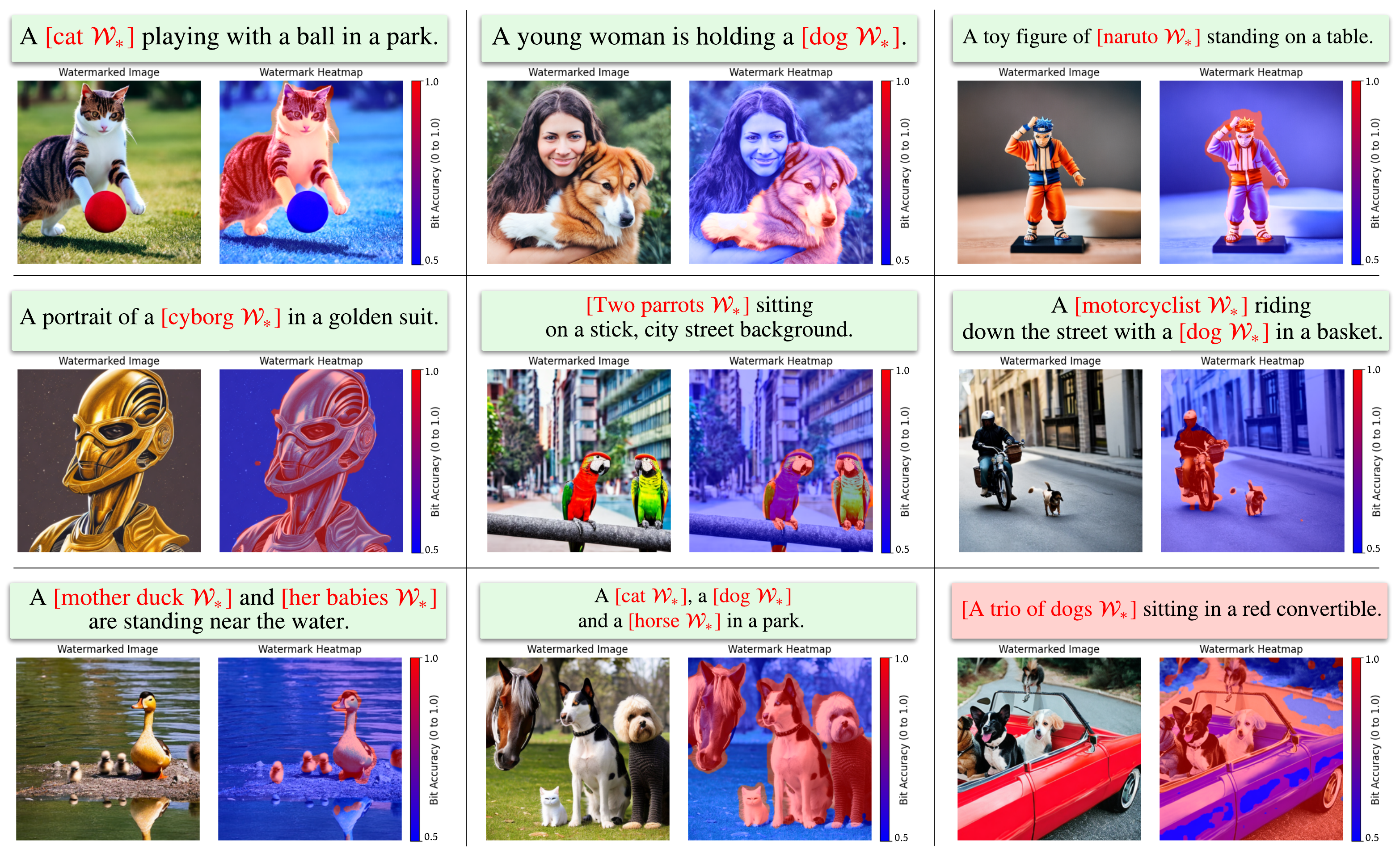}
    \caption{\textbf{Qualitative results and watermark heatmaps.} \emph{We present qualitative results of our watermarking approach while watermarking up to three objects within a single image. All the images are watermarked within the T2I generation pipeline. Recall from \cref{sec:obj_wat_method}, we control object-level watermarking directly from the text prompt $\mathcal{P}$. (Top row) presents single object watermarking along with the watermark heatmap for each generated image. We see high bit accuracy within each object selected for watermarking and 0 outside the object. (Second row, third row) presents our qualitative results for two and more than three objects, respectively. We see that our method can accurately retrieve watermarked objects with high bit accuracy. We see (bottom row, last column) that when attention maps are imperfectly defined, the watermark is not confined within the objects, but high bit accuracy is still achieved within the image.}}
    \label{fig:obj_qualitative}
\end{figure*}

\subsection{Key results on Object Watermarking} \label{sec:obj-watermarking}

\begin{figure}[!t]
    \centering
    \includegraphics[width=\linewidth]{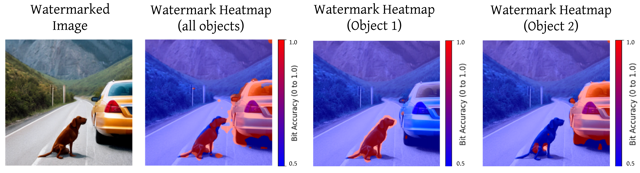}
    \caption{\textbf{Multiple Object watermarking heatmap breakdown.} \emph{Our goal is to embed watermark in specific regions of the image while not corrupting the entire image. In the above example, a user can choose to watermark either the dog or the car or both. Our method provides the versatility to choose to watermark any object(s) in an image while preserving other regions.}}
    \label{tab:heatmap-breakdown}
\end{figure}

\noindent {\bf Object Watermarking and Identification:} We present the effectiveness of object-level watermarking, as illustrated in \cref{fig:obj_qualitative}, where we apply watermarks to up to three distinct objects within a single image, mentioned in algorithm \cref{alg:UAC}. Using the watermark detector \( D_w \), and without the need for any additional information such as segmentation masks, heatmaps are generated by evaluating the bit accuracy of small patches, each covering approximately $10\%$ of the image area. Each patch's bit accuracy score is determined by the detector, and these scores are aggregated across all patches to produce a comprehensive heatmap. In this map, a score of $1$ represents a bit accuracy of $100\%$, indicating high fidelity in watermark retrieval, while a score of $0.5$ indicates retrieved watermark key does not match $m$ indicating no watermark. When watermarking a single object, the resulting heatmap shows precise retrieval of the watermark. For images with two or three watermarked objects, the detector reliably identifies each watermark, and accurate retrieval is observed. In cases involving four or more objects (bottom-right of \cref{fig:obj_qualitative}), the heatmap confirms that the watermark is detected across most of the image, likely due to the accuracy of attention maps during inference. For images containing multiple objects, separate heatmaps can be generated for each object to enhance clarity in visualization. The overall heatmap presented in \cref{fig:obj_qualitative} is an aggregate of individual heatmaps detailed in \cref{tab:heatmap-breakdown}. Further technical details on the heatmap generation process are available in the supplement (section B).

\input{sec/tables/table4_object_watermarking}

\noindent {\bf Stress Tests: Object Size and Multiple Object Watermarking:} We evaluate the sensitivity of the watermark detection to small object sizes and multiple objects. As shown in \cref{tab:obj_quantitative}, even with a $40\%$ crop of the object, the detector achieves $89\%$ bit accuracy. This accuracy remains robust to transformations. We also test watermark detection across multiple objects, with and without overlap. For non-overlapping objects, detection remains above $90\%$ without attacks and above $89\%$ under attacks. However, performance decreases when the overlap exceeds $40\%$, particularly in the case of multiple overlapping objects. Our qualitative results indicate that single-object watermarking is not affected by object overlap, but bit accuracy decreases with significant overlap due to multiple layers of interference.

{\bf Summary.} Object-level watermarking achieves high detection accuracy across multiple objects and varying sizes, with robust performance even under transformations, although overlap between objects during multi-object watermarking reduces bit accuracy.


\subsection{Applications: Integration with LDM Pipelines and Textual Inversion Compatibility}

\begin{figure}[!t]
    \centering
    \includegraphics[height=0.9\linewidth]{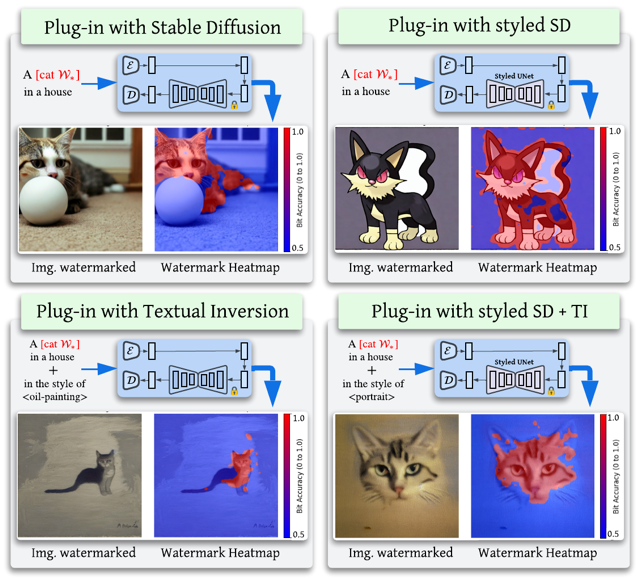}
    \vspace{-0.2in}
    \caption{\textbf{Plug-and-Play ability of our method.} \emph{We present our method’s ability to be plugged into any combination of personalized T2I model. Above image shows four such combinations where we use Textual Inversion style tokens and styled T2I pipelines can be seen. It can be observed that the object-level control and watermarking ability of our method is preserved across all these pipelines.} }
    \label{fig:plug-and-play}
    \vspace{-0.2in}
\end{figure}

We demonstrate the versatility of our method by integrating the plug-in watermark token $\mathcal{W}_*$ across diverse text-to-image (T2I) generation pipelines, assessing both watermark robustness and imperceptibility. The schematic in \cref{fig:method} illustrates this integration process, where $\mathcal{W}_*$ is loaded into the text encoder of different LDMs, including pipelines that employ personalized or fine-tuned diffusion models. Additionally, $\mathcal{W_*}$ can be combined with Textual Inversion tokens, as shown in qualitative results in \cref{fig:obj_qualitative}. Our method achieves a high PSNR of $35$ dB and bit accuracy above $92\%$, outperforming the Stable Signature baseline \cite{10377226} in robustness while maintaining watermark invisibility and resistance to attacks.

\subsection{Ablation Studies}
\label{sec:abl}
\begin{figure}[!t]
    \centering
    \includegraphics[width=\linewidth]{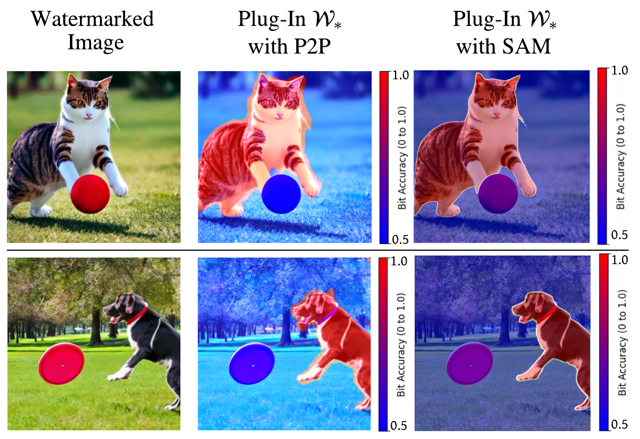}
    \caption{\textbf{Watermark heatmap enhancement using SAM.} \emph{Our watermarking is embedded into the generation pipeline using Attention maps. We test the performance of our watermark detection by plugging our method into P2P and SAM.}}
    \label{fig:ours_sam}
\end{figure}

\noindent {\bf Training with SAM Segmentation Masks:} Our watermarking pipeline relies on attention maps generated by prompts, though these maps can sometimes lack precision (e.g., spillover beyond the intended object, seen in bottom-right scenario of \cref{fig:obj_qualitative}). This can pose challenges for applications like medical imaging, where precise watermark localization is crucial. In case of a need for such high precision localization of watermark target, our method is flexible to utilize segmentations masks within watermark generation and need not rely on post-generation localization. In this ablation study, we incorporate SAM (Segment-Anything) segmentation masks directly into our generation process as an alternative to attention maps, particularly in Step~$4$ of \cref{alg:UAC}. Our results (\cref{fig:ours_sam}) show that SAM masks significantly improve watermark localization accuracy.



\noindent {\bf Ablation on Number of bits for Watermarking.} We perform an ablation on number of bits that can be embedded into image by our method and present our results as a plot in the supplement (section C.). We observe that our method can watermark with $> 91 \%$ bit accuracy for $128$ bits under various attacks.


%% file: sec/tables/table1_watermark_main.tex
\newcommand{\cmark}{\textcolor{green}{\ding{51}}} 
\newcommand{\xmark}{\textcolor{red}{\ding{55}}} 
\definecolor{darkgreen}{HTML}{EBFFEA} 
\begin{table*}[!t]
\vskip 0.1in
\begin{center}
\setlength\tabcolsep{2pt}
\scalebox{0.7}{ 
\begin{tabular}{lccccccccccccccccc} 
\toprule
Method & I.W. & O.W. & L.P. & \# Params $\downarrow$ & \multicolumn{3}{c}{Imperceptibility} & \multicolumn{7}{c}{Robustness to basic attacks (BA):} & \multicolumn{3}{c}{Adversarial attacks (BA):} \\ 
\cmidrule(lr){6-8} \cmidrule(lr){9-15} \cmidrule(lr){16-18}
& & & & & PSNR $\uparrow$ & SSIM $\uparrow$ & FID $\downarrow$ & 
None $\uparrow$ & Brightness $\uparrow$ & Contrast $\uparrow$ & Blur $\uparrow$ & Crop $\uparrow$ & Rot. $\uparrow$ & JPEG $\uparrow$ &  SDEdit $\uparrow$ & WMAttacker $\uparrow$ & DiffPure $\uparrow$ \\ 
\midrule
\multicolumn{18}{l}{\underline{\textit{WikiArt (48 bits) - In-Generation Watermarking}}} \vspace{1mm} \\ 
Stable Sig. \citep{10377226} & \cmark & \xmark & \xmark & $10^5+$ & 31.57 & 0.88 & 24.71 & 0.99 & 0.93 & 0.87 & 0.78 & 0.79 & 0.70 & 0.55 & 0.58 & 0.53 & 0.52 \\ 
LaWa \citep{rezaei2024lawausinglatentspace} & \cmark & \xmark & \xmark & $10^5+$ & 32.52 & 0.93 & 18.23 & 0.99 & \textbf{0.99} & \underline{0.98} & 0.94 & 0.93 & 0.83 & 0.89 & 0.68 & 0.76 & 0.78 \\ 
TrustMark \citep{bui2023trustmarkuniversalwatermarkingarbitrary} & \xmark & \xmark & \xmark & $10^5+$ & \underline{39.90} & 0.97 & \underline{15.83} & 0.99 & 0.98 & 0.99 & 0.93 & 0.89 & 0.87 & 0.86 & 0.68 & 0.77 & 0.73 \\ 
RoSteALS \citep{bui2023rostealsrobuststeganographyusing} & \cmark & \xmark & \xmark & $10^5+$ & 32.68 & 0.88 & 16.63 & \underline{0.98} & 0.96 & 0.94 & 0.88 & 0.88 & 0.75 & 0.80 & 0.75 & 0.72 & 0.72 \\ 
AquaLoRA \citep{pmlr-v235-feng24k} & \cmark & \xmark & \xmark & $10^5+$ & 31.46 & 0.92 & 17.27 & 0.94 & 0.91 & 0.91 & 0.81 & 0.90 & 0.58 & 0.76 & 0.68 & 0.67 & 0.66 \\ 
WAM \citep{sander2024watermarklocalizedmessages} & \cmark & \cmark & \xmark & $10^5+$ & 36.46 & 0.97 & 16.27 & 0.97 & 0.93 & 0.92 & 0.84 & 0.92 & 0.76 & 0.84 & 0.72 & 0.71 & 0.72 \\ 
\rowcolor{darkgreen}\textbf{Ours + SD$_\text{style}$} & \cmark & \cmark & \cmark & \textbf{768} & 35.88 & 0.93 & 16.72 & 0.99 & 0.97 & 0.99 & 0.94 & \underline{0.97} & \underline{0.94} & \underline{0.93} & 0.81 & 0.83 & 0.84 \\ 
\rowcolor{darkgreen} \textbf{Ours + TI} & \cmark & \cmark & \cmark & \textbf{768} & 36.89 & \underline{0.94} & 15.98 & 0.99 & 0.96 & 0.99 & \underline{0.95} & 0.98 & 0.95 & 0.92 & \underline{0.82} & \underline{0.84} & \underline{0.86} \\ 
\rowcolor{darkgreen} \textbf{Ours + SD} & \cmark & \cmark & \cmark & \textbf{768} & \textbf{39.92} & \textbf{0.97} & \textbf{14.89} & \textbf{0.99} & \underline{0.98} & \textbf{0.99} & \textbf{0.97} & \textbf{0.98} & \textbf{0.95} & \textbf{0.95} & \textbf{0.85} & \textbf{0.87} & \textbf{0.88} \\ 
\midrule
\multicolumn{18}{l}{\underline{\textit{MS-COCO (48 bits) - In-Generation Watermarking}}} \vspace{1mm} \\ 
Stable Sig. \citep{10377226} & \cmark & \xmark & \xmark & $10^5+$ & 31.93 & 0.88 & 24.74 & 0.99 & 0.94 & 0.89 & 0.81 & 0.82 & 0.72 & 0.59 & 0.62 & 0.63 & 0.57 \\ 
LaWa \citep{rezaei2024lawausinglatentspace} & \cmark & \xmark & \xmark & $10^5+$ & 33.53 & 0.95 & 17.45 & 0.99 & \textbf{0.99} & \underline{0.98} & 0.94 & 0.93 & 0.86 & 0.90 & 0.71 & 0.79 & 0.79 \\ 
TrustMark \citep{bui2023trustmarkuniversalwatermarkingarbitrary} & \xmark & \xmark & \xmark & $10^5+$ & \textbf{40.98} & \textbf{0.98} & \underline{14.89} & 0.99 & 0.98 & 0.99 & \underline{0.95} & 0.91 & 0.89 & 0.88 & 0.69 & 0.78 & 0.74 \\ 

RoSteALS \citep{bui2023rostealsrobuststeganographyusing} & \cmark & \xmark & \xmark & $10^5+$ & 32.68 & 0.88 & 16.63 & 0.99 & 0.98 & 0.93 & 0.89 & 0.87 & 0.78 & 0.81 & 0.76 & 0.72 & 0.75 \\ 
AquaLoRA \citep{pmlr-v235-feng24k} & \cmark & \xmark & \xmark & $10^5+$ & 31.46 & 0.92 & 17.27 & \underline{0.95} & 0.91 & 0.90 & 0.80 & 0.92 & 0.68 & 0.78 & 0.67 & 0.70 & 0.68 \\ 
WAM \citep{sander2024watermarklocalizedmessages} & \cmark & \cmark & \xmark & $10^5+$ & 36.98 & 0.97 & 16.85 & 0.98 & 0.94 & 0.92 & 0.86 & 0.94 & 0.77 & 0.86 & 0.73 & 0.73 & 0.72 \\ 
\rowcolor{darkgreen}\textbf{Ours + SD$_\text{style}$} & \cmark & \cmark & \cmark & \textbf{768} & {36.99} & {0.93} & {16.72} & {0.99} & {0.98} & {0.99} & {0.94} & \underline{0.97} & {0.94} & {0.93} & {0.81} & {0.83} & {0.84} \\ 
\rowcolor{darkgreen} \textbf{Ours + TI} & \cmark & \cmark & \cmark & \textbf{768} & {36.89} & {0.94} & {15.23} & {0.99} & {0.96} & {0.99} & \underline{0.95} & {0.98} & \underline{0.96} & \underline{0.94} & \underline{0.83} & \underline{0.84} & \underline{0.86} \\ 
\rowcolor{darkgreen} \textbf{Ours + SD} & \cmark & \cmark & \cmark & \textbf{768} & \underline{40.92} & \underline{0.97} & \textbf{14.83} & \textbf{0.99} & \underline{0.98} & \textbf{0.99} & \textbf{0.97} & \textbf{0.98} & \textbf{0.97} & \textbf{0.96} & \textbf{0.86} & \textbf{0.88} & \textbf{0.89} \\
\bottomrule
\end{tabular}
}
\end{center}
\vskip -0.2in
\caption{\textbf{Comparison to watermark baselines.} \emph{(I.W.: In-generation Watermarking, O.W.: Object-level Watermarking, L.P.: Less than $10^5$ parameters, BA: Bit Accuracy) We compare our method several baselines. In addition to watermark invisibility and robustness of watermarking, we present the number of parameters used for training. We see that our method uses $10^5 \times$ lower parameters. We see that early integration for watermarking improves robustness to attacks, we see a consistent trend of this improvements in basic image processing attacks and adversarial attacks. We also present the performance of our method in the presence of personalization using fine-tuned UNet and Textual Inversion. From the above table we see that our method can be plugged into various LDM pipelines. More details on specific implementation of attacks can be found in the supplement.}}
\label{tab:modified-table-main}
\end{table*}

%% file: sec/tables/table4_object_watermarking.tex
\begin{table}[!t]
    \centering
    \normalsize  
    \setlength{\tabcolsep}{5pt}  

    \resizebox{\linewidth}{!}{
        \begin{tabular}{l c c c c c c}
        \toprule
        \multicolumn{1}{c}{} & \multicolumn{6}{c}{\textbf{Bit Accuracy}} \\ 
        \cmidrule(lr){2-7}
        \textbf{} & \textbf{None} & \textbf{Bright.} & \textbf{Cont.} & \textbf{Blur} & \textbf{Rot.} & \textbf{JPEG} \\ 
        \midrule
        
        \multicolumn{7}{l}{\textit{\underline{Single object}}} \\  
        Segment + white bg           & $0.99$  & $0.97$ & $0.96$ &  $0.97$  &  $0.96$  &  $0.97$  \\ 
        Segment + style bg           & $0.95$  & $0.92$ & $0.90$ &  $0.96$  &  $0.94$  &  $0.93$  \\ 
        Crop object (0.8 $\times$ size) & $0.96$  & $0.94$ & $0.92$ & $0.95$  & $0.92$  & $0.93$  \\
        Crop object (0.5 $\times$ size) & $0.92$  & $0.91$ & $0.90$ & $0.91$  & $0.92$  & $0.90$  \\
        Crop object (0.4 $\times$ size) & $0.90$  & $0.89$ & $0.88$ & $0.89$  & $0.89$  & $0.90$  \\ 
        \midrule

        \multicolumn{7}{l}{\textit{Multiple objects}} \\  
        1 object                     & $0.99$  & $0.97$ & $0.96$ &  $0.97$  &  $0.95$  &  $0.98$ \\ 
        2 objects (no overlap)        & $0.94$  & $0.93$ & $0.95$ & $0.95$  & $0.94$  & $0.96$  \\ 
        3 objects (no overlap)        & $0.90$  & $0.89$ & $0.90$ & $0.90$  & $0.90$  & $0.99$  \\ 
        2 objects (overlap $\geq 40\%$) & $0.79$  & $0.76$ & $0.70$ & $0.80$  & $0.74$  & $0.74$  \\
        \bottomrule
    \end{tabular}}
    \caption{
        \textbf{Object-level watermarking robustness performance with attacks.} \emph{We evaluate the performance of our method to perform object-level watermarking in the presence of attacks. We clearly see that our method achieves robust watermarking performance on rotate and crop attacks for different crops performed within watermarked objects. We present our results while watermarking single and up to three objects.} 
        \vspace*{-0.2cm}
    }\label{tab:obj_quantitative}
     
\end{table}

%% file: sec/6_conclusion.tex
\section{Limitations}

As our method performs in-generation watermarking. For localizing the watermark on an object, the method relies on Cross Attention maps for each token in the input prompt $\mathcal{P}$. Hence, the method relies strongly on the accuracy of these cross attention maps. Our ablation studies conducted on using the segmentation masks from SAM enhance the localization of watermark on top of the object. However, relying on an external segmentation model such as SAM could be undesirable and overly relying on Cross Attention maps could be a limitation when the attention maps are not properly defined.

\section{Conclusions}

In this paper, we propose a novel in-generation watermarking technique to integrate watermarking into the latent within the denoising process of T2I generation. Our watermarking technique provides watermarking control directly from text and fine-tunes token embeddings of a single token. Our method contributes to a novel application of object-level watermarking within T2I generation. We show that our early watermarking technique shows improvements in watermarking robustness across several post generation attacks. Our method aims to motivate future research towards training-free watermarking, controllable watermarking with any T2I generation pipeline.

%% file: sec/X_suppl.tex
\onecolumn
\appendix
\thispagestyle{empty}

\maketitlesupplementary

\section{Post-generation vs. In-generation Watermarking: Limitations and Advances}

In \textit{post-generation watermarking}, also known as post-hoc watermarking, watermarks are injected into images after generation \citep{an2024benchmarking}. This approach, while straightforward, incurs additional computational overhead and is vulnerable to circumvention. For example, in cases of model leakage, attackers can easily detect and bypass the postprocessing module \cite{wen2024tree}. Additionally, post-diffusion methods often result in poorer image quality, introducing artifacts, and their separable nature makes them easily removable in open-source models, such as by commenting out a single line of code in Stable Diffusion code base \cite{10377226}.

In contrast, \textit{in-generation watermarking} integrates the watermarking process into the image generation pipeline, improving stealthiness and computational efficiency. This approach embeds watermarks directly into generated images without requiring separate post-hoc steps. It is also less susceptible to diffusion denoising or removal via simple modifications.

The closest related work is ~\cite{min2024watermark}, which also embeds watermarking bits during generation. However, their method requires retraining a UNet layer, limiting its plug-and-play capability, as discussed in Section 2 and Figure 4. Our approach, by comparison, is more convenient, embedding watermarks directly within the text prompt as a watermarking token.

Focusing on object watermarking, our method differs from~\cite{zhao2024ssyncoa} and~\cite{sander2024watermark} in several key aspects. While these methods rely on segmentation maps for supervised watermarking of specific objects in pre-existing images, ours is unsupervised, leveraging attention maps generated automatically during text-to-image generation. Unlike post-hoc methods, which watermark only pre-existing images, our approach simultaneously performs text-to-image generation and watermarking. This enables compatibility with techniques like textual inversion and ensures robustness against attacks, overcoming significant limitations of post-hoc methods.

\begin{table}[!t]
    \centering

    \scalebox{0.55}{  
    \begin{tabular}{lcccccccc}
        \toprule
        \multirow{2}{*}{Method} & \multicolumn{3}{c}{Imperceptibility} & \multicolumn{5}{c}{Robustness to Basic Attacks (BA):} \\ 
        \cmidrule(lr){2-4} \cmidrule(lr){5-9}
        & PSNR $\uparrow$ & SSIM $\uparrow$ & FID $\downarrow$ & None $\uparrow$ & Bright. $\uparrow$ & Contrast $\uparrow$ & Blur $\uparrow$ & JPEG $\uparrow$ \\ 
        \midrule
        Dct-Dwt & 39.50 & 0.97 & 15.93 & 0.96 & 0.89 & 0.91 & 0.90 & 0.55 \\ 
        SSL Watermark & 31.50 & 0.86 & 21.82 & 0.95 & 0.91 & 0.84 & 0.88 & 0.55 \\ 
        HiDDeN & 31.57 & 0.88 & 22.67 & 0.99 & 0.93 & 0.88 & 0.80 & 0.88 \\
        \textbf{Ours with \cite{zhu2018hiddenhidingdatadeep}} & \textbf{37.92} & \textbf{0.95} & \textbf{15.83} & \textbf{0.99} & \textbf{0.99} & \textbf{0.97} & \textbf{0.96} & \textbf{0.96} \\
        \textbf{Ours with \cite{pmlr-v235-feng24k}} & \textbf{40.92} & \textbf{0.97} & \textbf{14.83} & \textbf{0.99} & \textbf{0.98} & \textbf{0.99} & \textbf{0.97} & \textbf{0.96} \\
        \bottomrule
    \end{tabular}
    }
    \caption{\textbf{Performance comparison of our method with post generation watermarking methods.} 
    We evaluate imperceptibility and robustness against basic attacks. Higher PSNR and SSIM, and lower FID, indicate better imperceptibility. Higher robustness scores indicate greater resistance to attacks.}
    \label{tab:watermarking_performance}
\end{table}

\section{Algorithm for Watermark Heatmap Generation}
We present an algorithm for generating the heatmaps depicted in Figure 4 of the main paper. The process involves iterating a defined patch across the image to construct the heatmap. Details regarding the minimum allowable size of the patch can be found in Section 4.4. The patch represents the smallest region of the image that can reliably extract bits using a message detector with high confidence; for further explanation, refer to Section 4 in the main paper. Notably, the detector employed is an off-the-shelf solution. The resulting heatmap is normalized to a range of $[0, 1]$, where a value of $0$ indicates regions with $0\%$ bit accuracy, and a value of $1$ corresponds to regions achieving $100\%$ bit accuracy. Lastly, we apply Gaussian blur that preserves the edges and boundaries better than other uniform blurring filters, which is important for maintaining the structure of the heatmap.

\begin{algorithm}[!h]
    \caption{Watermark Heatmap Generation} \label{alg:heatmap}
    \begin{algorithmic}[1]
        \Statex \textbf{Input:} Watermarked image $\mathbf{I}$; Watermark Detector $\mathcal{D}_w$; Ground Truth Key $m$.
        \State Select a patch dimensions $h \times w$.
        \State Divide $\mathbf{I}$ into overlapping patches $\{P_{ij}\}$, where $P_{ij}$ is the patch at location $(i, j)$ of size $h \times w$.
        \State Initialize a heatmap matrix $\mathbf{H}$ of zeros with the same spatial resolution as $\mathbf{I}$.
        \For{each patch $P_{ij}$ in $\mathbf{I}$}
            \State Extract watermark string from the patch: 
            \[
            m'_{ij} = \mathcal{D}_w(P_{ij})
            \]
            \State Compute the bit accuracy for the patch as (where $|m|$ is the length of the bit string):
            \[
            A_{ij} = \frac{1}{|m|} \sum_{k=1}^{|m|}[m_k = m'_{ij,k}]
            \]
            \State Assign the bit accuracy value to the center pixel of the patch in the heatmap $\mathcal{H}$:
            \[
            \mathbf{H}(i,j) \gets A_{ij}
            \]
        \EndFor
        \State Normalize the heatmap values to the range $[0, 1]$:
        \[
        \mathbf{H}_{\text{norm}} = \frac{\mathbf{H} - \min(\mathbf{H})}{\max(\mathbf{H}) - \min(\mathbf{H})}.
        \]
        \State Apply a smoothing filter (Gaussian blur) to $\mathbf{H}_{\text{norm}}$:
        \[
        \mathbf{H}_{\text{smooth}} = \text{GaussianBlur}(\mathbf{H}_{\text{norm}}, \sigma).
        \]
        \State Generate a heatmap image by overlaying $\mathbf{H}_{\text{smooth}}$ on the original watermarked image $\mathbf{I}$.
        \Statex \textbf{Output:} Heatmap image highlighting regions with watermark accuracy.
    \end{algorithmic}
\end{algorithm}

\section{Number of Bits v/s Bit Accuracy}

We perform an ablation study on the number of bits that can be embedded into images by our method, without sacrificing on bit accuracy. Methods such as \cite{bui2023rostealsrobuststeganographyusing} can embed upto $100$ bits during watermarking. Extending the effort to push the benchmark for number of bits, we show our results on watermarking upto 128 bits.

\begin{figure}[!h]
    \centering
    \includegraphics[width=1\linewidth]{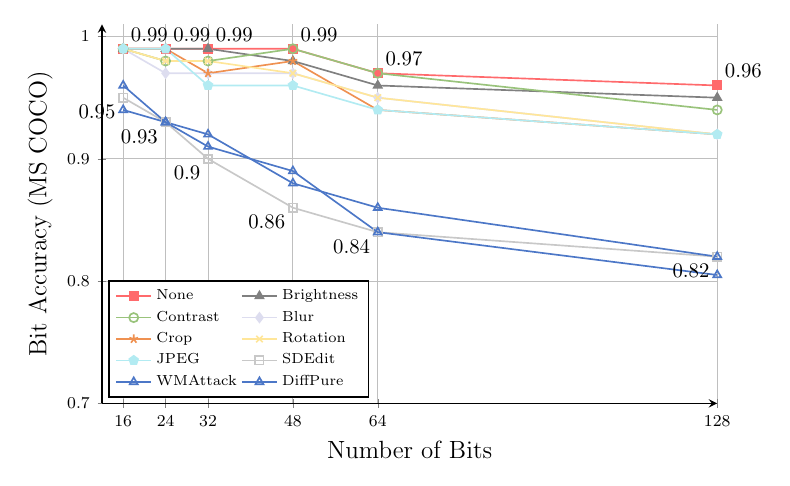}
    \caption{\textbf{Number of bits v/s Bit Accuracy:} It can be seen from the above plot that our method can embed 128 bits without loss in bit accuracy.  We observe a bit accuracy of over 90\% with 128 bits.}
    \label{fig:enter-label}
\end{figure}

\section{Comparison to methods that use TPR as evaluation metric}

We report True Positive Rate of our watermark detection in comparison with baselines including \cite{wen2024tree, lei2024conceptwmdiffusionmodelwatermark}. We threshold the FPR to 0.1\% and report TPR:0.1\%

\begin{table}[!h]
    \centering

    \scalebox{0.55}{  
    \begin{tabular}{lcccccccc}
        \toprule
        \multirow{2}{*}{Method} & \multicolumn{3}{c}{Imperceptibility} & \multicolumn{5}{c}{Robustness to Basic Attacks (TPR):} \\ 
        \cmidrule(lr){2-4} \cmidrule(lr){5-9}
        & PSNR $\uparrow$ & SSIM $\uparrow$ & FID $\downarrow$ & None $\uparrow$ & Bright. $\uparrow$ & Contrast $\uparrow$ & Blur $\uparrow$ & JPEG $\uparrow$ \\ 
        \midrule
        TreeRings & 33.75 & 0.91 & 18.93 & \textbf{1.00} & 0.91 & 0.90 & 0.92 & 0.89 \\ 
        ConceptWM & 32.89 & 0.89 & 24.58 & 0.98 & 0.90 & 0.84 & 0.83 & 0.85 \\ 
        \textbf{Ours} & \textbf{40.92} & \textbf{0.97} & \textbf{14.83} & 0.99 & \textbf{0.99} & \textbf{0.98} & \textbf{0.99} & \textbf{0.97} \\
        \bottomrule
    \end{tabular}
    }
    \caption{\textbf{Performance comparison of our method with methods that use TPR as evaluation metric.} 
    We evaluate imperceptibility and robustness against basic attacks. Higher PSNR and SSIM, and lower FID, indicate better imperceptibility. Higher robustness scores indicate greater resistance to attacks.}
    \label{tab:watermarking_performance}
\end{table}

\section{Empirical study to find the optimal timestep for watermarking}

As discussed in the ablation studies section of our paper, we perform empiracal study to find the optimal timestep $\tau^*$ that provides an balance between the trade off between watermark invisibility and bit accuracy.

\begin{figure}[!h]
    \centering
    \includegraphics[width=1\linewidth]{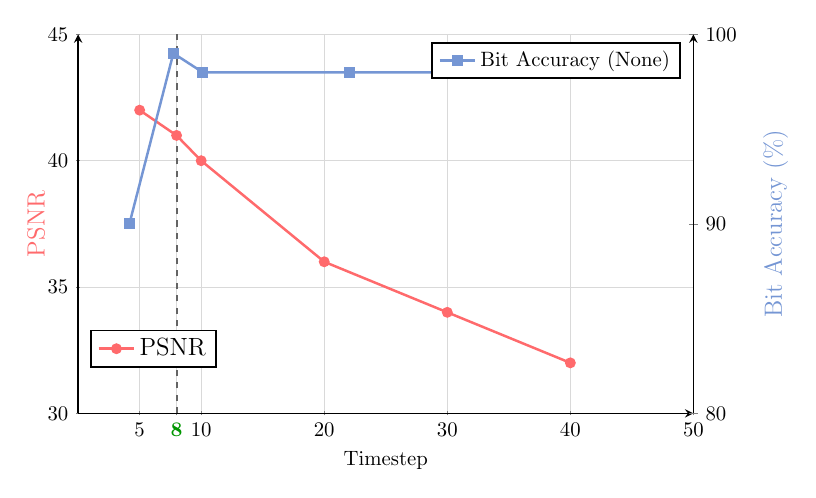}
    \caption{\textbf{Optimal sub-range of noise timesteps for invisible watermarking.} Image quality and watermark robustness are crucial evaluation metrics for watermarking techniques. Our method integrates watermarking into the core denoising module of a T2I pipeline. We study the evaluation metrics for various noise timesteps, and we find that to maintain invisible watermarking, the timesteps closest to the LDM encoder during forward process are optimal. We test this finding for multiple runs to utilize this finding into our watermarking method.}
    \label{fig:enter-label}
\end{figure}

\section{Details on Training and Inference time for watermarking}

As discussed in our method adds $\tau^*$ timesteps of noise during the forward processing and performs denoising to train $\bm{\mathcal{W}_*}$ token embeddings. We use 2000 images from MS COCO dataset for training and the entire training process takes about $2$ GPU hours when benchmarked on NVIDIA RTX A6000 GPU.

During inference, the time overhead is minimal, the significant storage savings (a $10^5 \times$ reduction in parameters) are noteworthy. Benchmarking on an RTX A6000, processing a single image (averaged over 1000 images) shows that for a T2I model, processing 50 diffusion steps takes 5.2 seconds without watermarking and 5.4 seconds with a minimal
0.2-second overhead due to the watermarking token.


\begin{figure*}[!h]
    \centering
    \begin{minipage}{\linewidth}
        \centering
        \includegraphics[width=\linewidth]{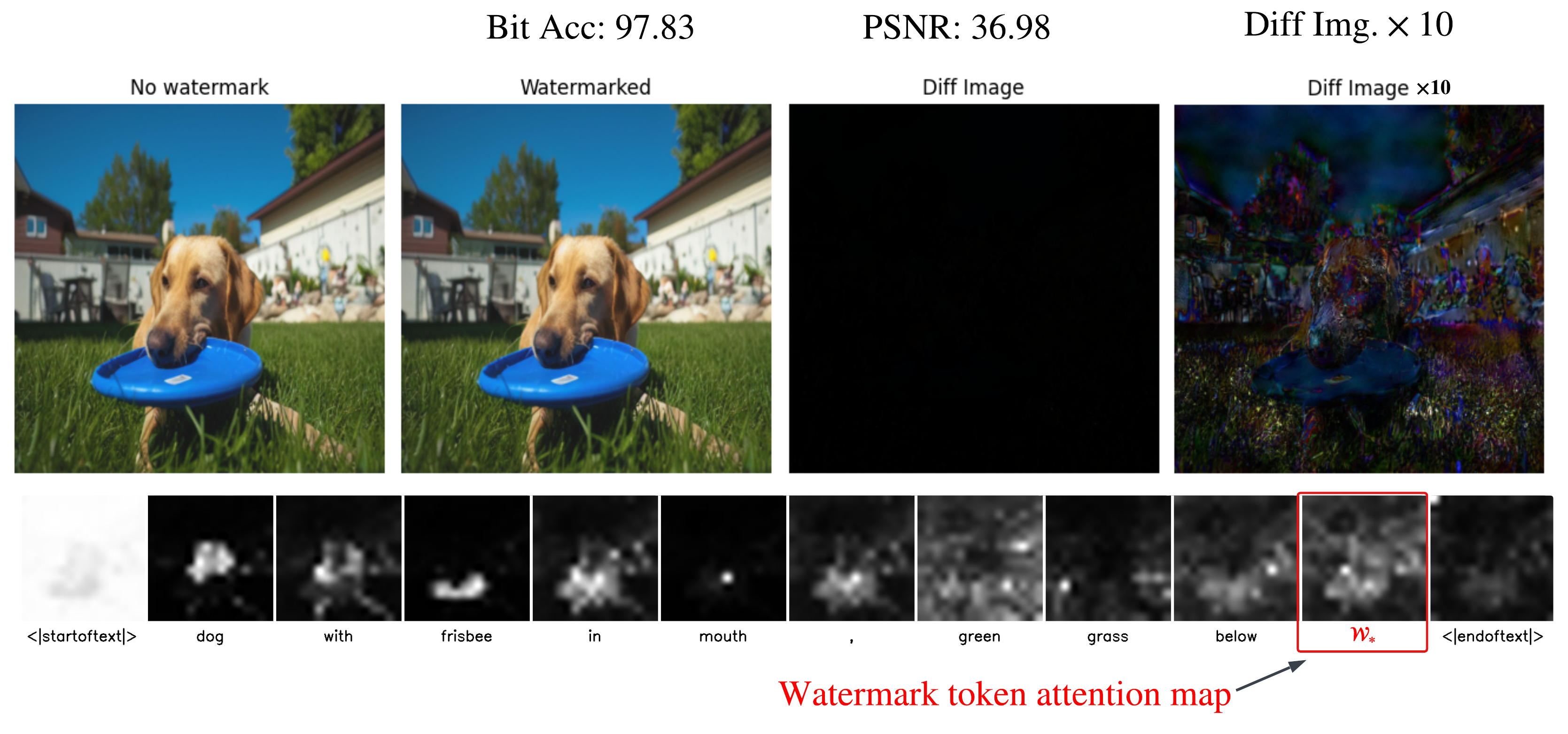}
    \end{minipage}
    
    \vspace{2mm} 
    
    \begin{minipage}{\linewidth}
        \centering
        \includegraphics[width=\linewidth]{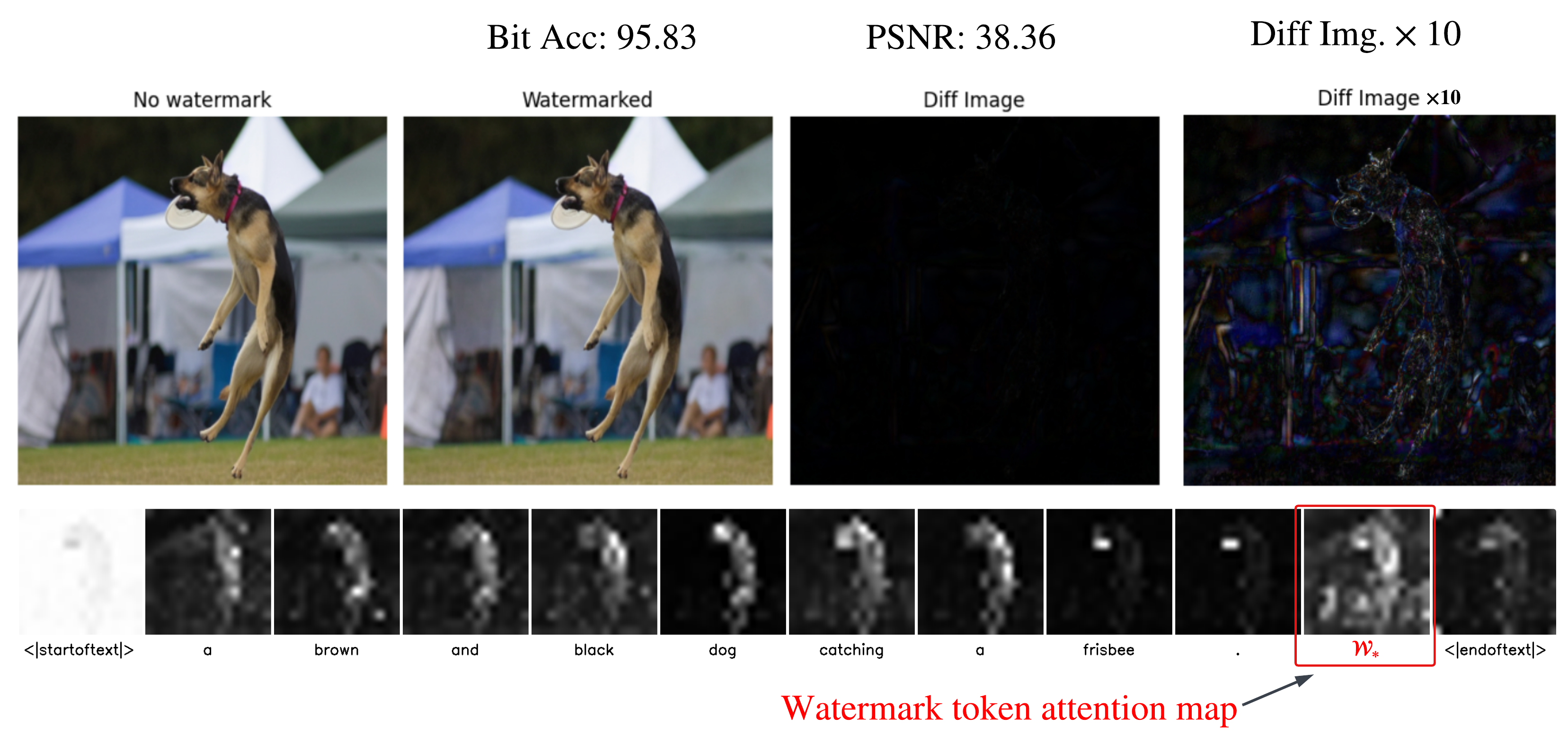}
    \end{minipage}

    \caption{\textbf{Watermark Invisibility and $\mathcal{W}_*$ Attention Map} \emph{We provide the attention map of our watermarking token $\mathcal{W}_*$. Our method achieves a very high PSNR of 38.36 while maintaining a bit accuracy of over 95\%. The attention map of $\mathcal{W}_*$ does not interfere with the attention maps of other tokens, preserving overall image quality.}}
    \label{fig:watermark_attn_map}
\end{figure*}

\clearpage

\begin{figure*}[!h]
    \centering
    \begin{minipage}{\linewidth}
        \centering
        \includegraphics[width=\linewidth]{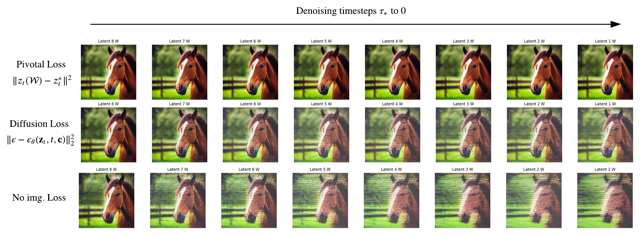}
    \end{minipage}
    
    \vspace{2mm} 
    
    \begin{minipage}{\linewidth}
        \centering
        \includegraphics[width=\linewidth]{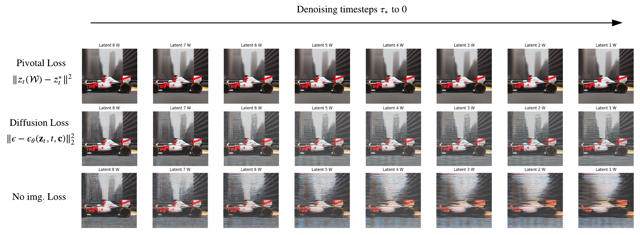}
    \end{minipage}

    \caption{\textbf{Diffusion Loss v/s Latent Loss} \emph{We provide qualitative results for our choice of latent loss in our method. This figure shows decoded latents at each step of denoising from $\tau_*$ to $0$ within T2I generation. The first row corresponds to our method, which achieves the best image quality while watermarking. The second row represents training using Diffusion Loss, while the last row shows image quality when $\mathcal{W}_*$ is trained only on watermarking loss without any image loss applied.}}
    \label{fig:diffusion_pivotal_loss}
\end{figure*}

\clearpage
\begin{figure*}
\centering
    \scriptsize
    \newcommand{\imwidth}{0.165\textwidth}
        \setlength{\tabcolsep}{0pt}
        \begin{tabular}{c@{\hskip 2pt}cccc}
        \toprule
        Original & AquaLoRA & Stable Signature & RoSteALS & Ours (Object) \\
        \midrule

        \multicolumn{5}{c}{\textit{Prompt: An old-fashioned \textcolor{red}{[drink $\mathcal{W}_*$]} next to a napkin}} \vspace{1mm} \\
        
        \includegraphics[width=\imwidth]{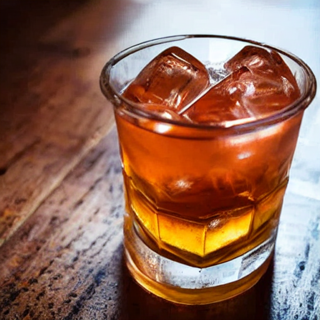} &
        \includegraphics[width=\imwidth]{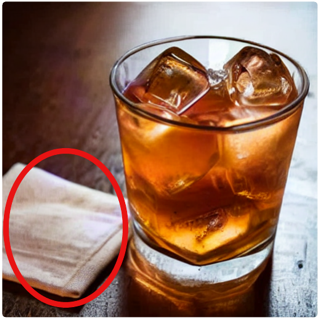} &
        \includegraphics[width=\imwidth]{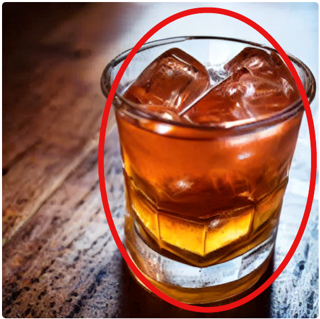} &
        \includegraphics[width=\imwidth]{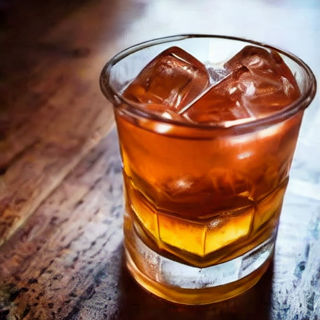} &
        \includegraphics[width=\imwidth]{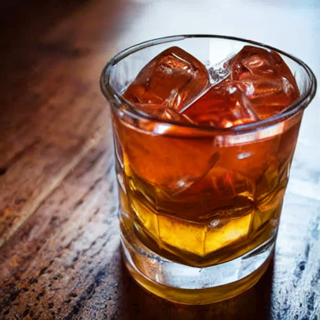} \\

         &
        \includegraphics[width=\imwidth]{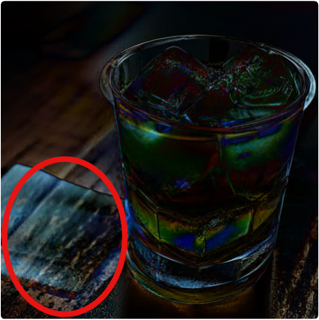} &
        \includegraphics[width=\imwidth]{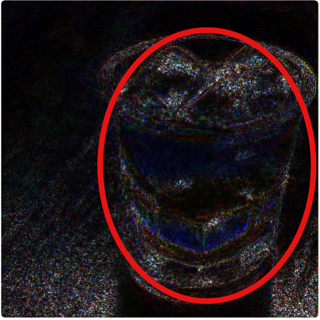} &
        \includegraphics[width=\imwidth]{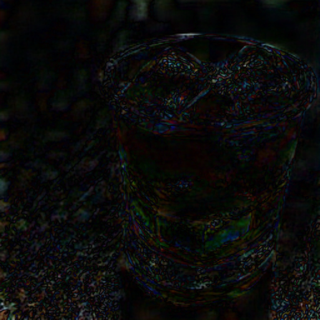} &
        \includegraphics[width=\imwidth]{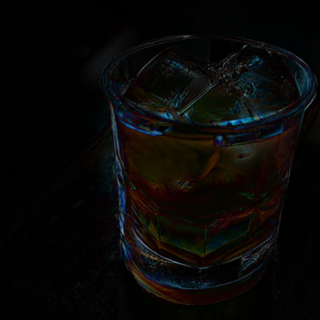} \\
        \midrule

        \multicolumn{5}{c}{\textit{Prompt: A toy car in front of a \textcolor{red}{[teddy bear $\mathcal{W}_*$]}}} \vspace{1mm} \\
        
        \includegraphics[width=\imwidth]{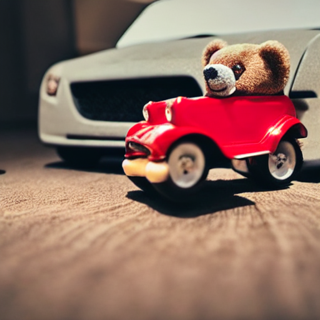} &
        \includegraphics[width=\imwidth]{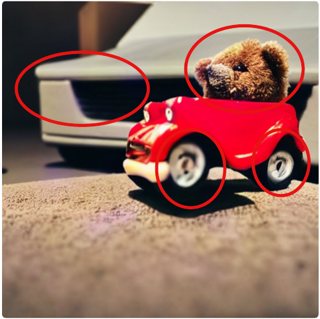} &
        \includegraphics[width=\imwidth]{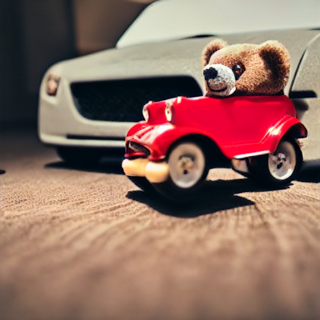} &
        \includegraphics[width=\imwidth]{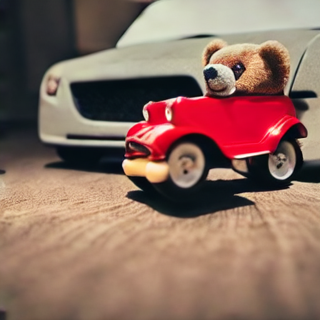} &
        \includegraphics[width=\imwidth]{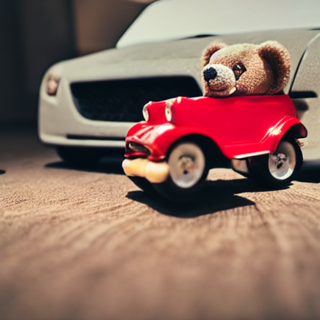} \\

         &
        \includegraphics[width=\imwidth]{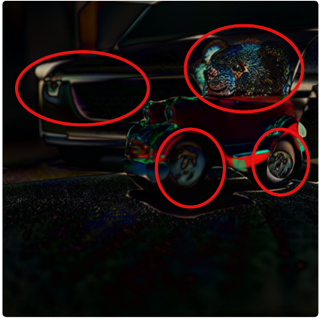} &
        \includegraphics[width=\imwidth]{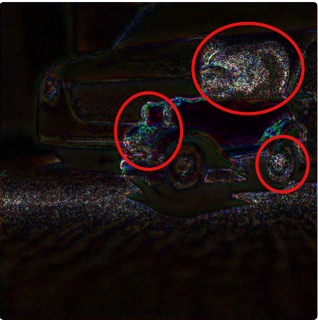} &
        \includegraphics[width=\imwidth]{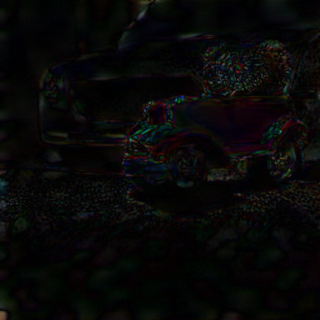} &
        \includegraphics[width=\imwidth]{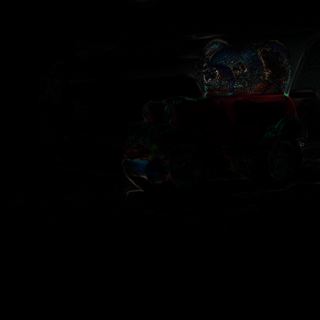} \\
        \midrule

        \multicolumn{5}{c}{\textit{Prompt: A \textcolor{red}{[woman $\mathcal{W}_*$]} running on a trail}} \vspace{1mm} \\
        
        \includegraphics[width=\imwidth]{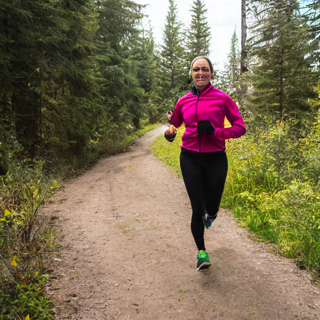} &
        \includegraphics[width=\imwidth]{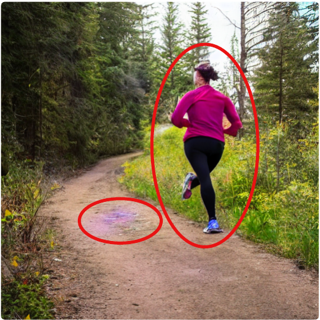} &
        \includegraphics[width=\imwidth]{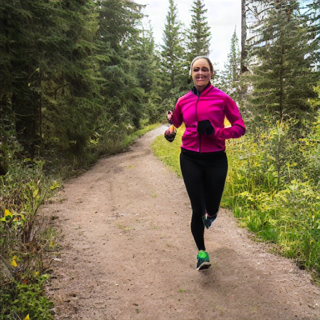} &
        \includegraphics[width=\imwidth]{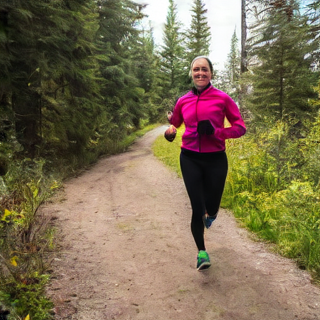} &
        \includegraphics[width=\imwidth]{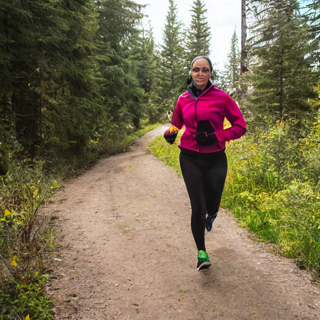} \\

         &
        \includegraphics[width=\imwidth]{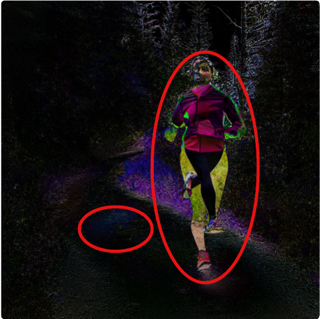} &
        \includegraphics[width=\imwidth]{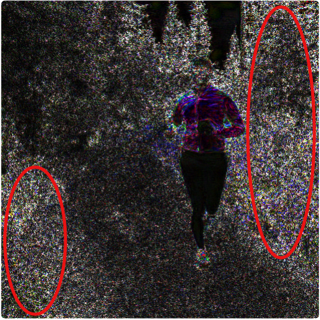} &
        \includegraphics[width=\imwidth]{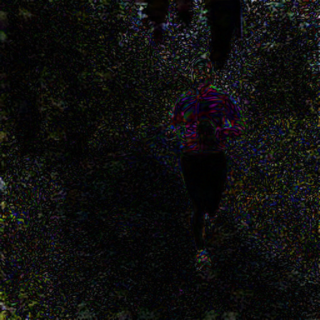} &
        \includegraphics[width=\imwidth]{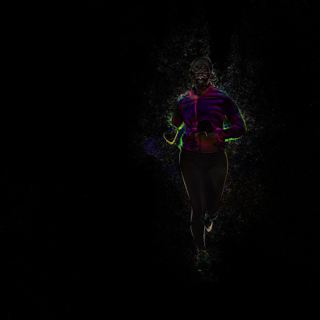} \\
        \midrule

        \end{tabular}
        \caption{\label{fig:supp-watermark-part2} \textbf{Qualitative results 1 for different watermarking methods on generated images} \emph{We provide several comparative examples with existing watermarking techniques for a qualitative analysis of watermark invisibility. The above images compare the performance of our method with AquaLoRA \cite{pmlr-v235-feng24k}, Stable Signature \cite{10377226} and RoSteALS \cite{bui2023rostealsrobuststeganographyusing}. We see that we surpass existing watermark methods in maintaining invisibility while watermarking an object within an image. In addition to the watermarked image, we also provide difference image below each watermarked image. Text in red denotes the object being watermarked by our method.} }
\end{figure*}

\clearpage
\begin{figure*}[ht]
\centering
    \scriptsize
    \newcommand{\imwidth}{0.165\textwidth}
        \setlength{\tabcolsep}{0pt}
        \begin{tabular}{c@{\hskip 2pt}cccc}
        \toprule
        Original & AquaLoRA & Stable Signature & RoSteALS & Ours (Object) \\
        \midrule

        \multicolumn{5}{c}{\textit{Prompt: The \textcolor{red}{[Great Wall $\mathcal{W}_*$]}}} \vspace{1mm} \\
        
        \includegraphics[width=\imwidth]{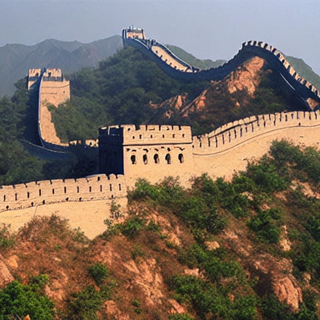} &
        \includegraphics[width=\imwidth]{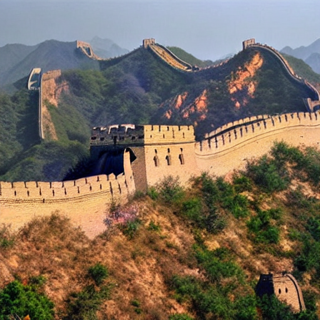} &
        \includegraphics[width=\imwidth]{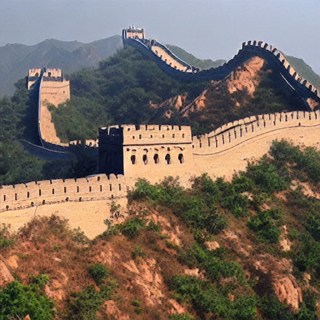} &
        \includegraphics[width=\imwidth]{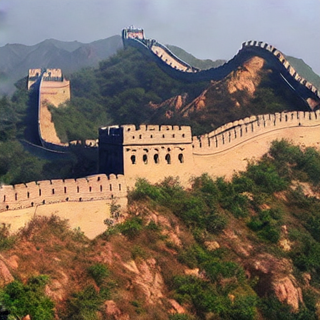} &
        \includegraphics[width=\imwidth]{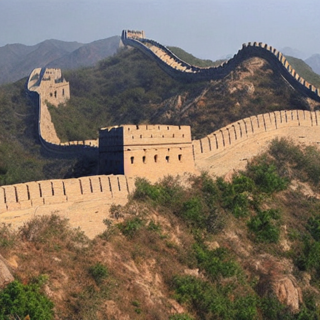} \\

         &
        \includegraphics[width=\imwidth]{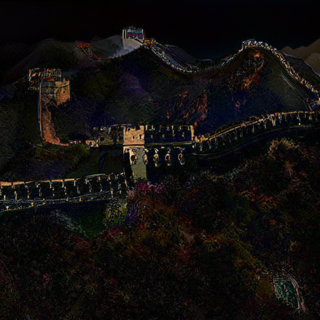} &
        \includegraphics[width=\imwidth]{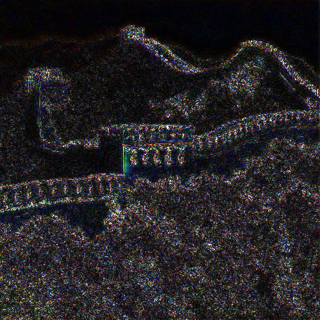} &
        \includegraphics[width=\imwidth]{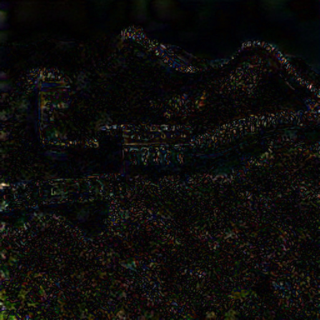} &
        \includegraphics[width=\imwidth]{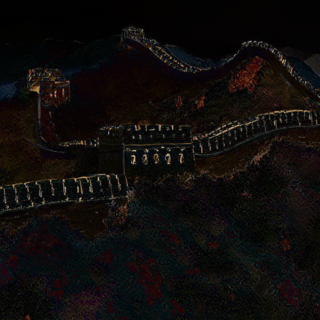} \\
        \midrule

        \multicolumn{5}{c}{\textit{Prompt: Brown white and black white \textcolor{red}{[guinea pigs$\mathcal{W}_*$]} eating parsley handed to them}} \vspace{1mm} \\
        
        \includegraphics[width=\imwidth]{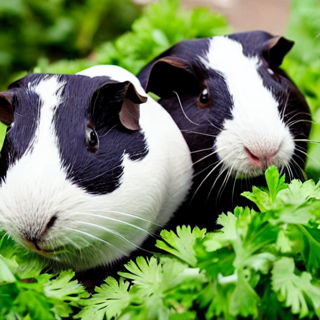} &
        \includegraphics[width=\imwidth]{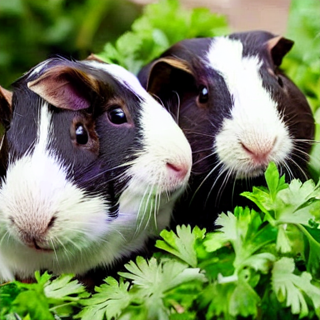} &
        \includegraphics[width=\imwidth]{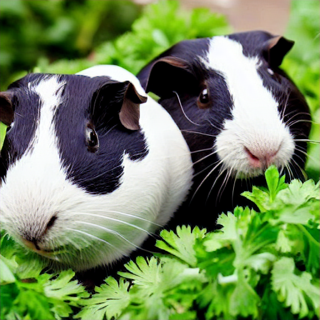} &
        \includegraphics[width=\imwidth]{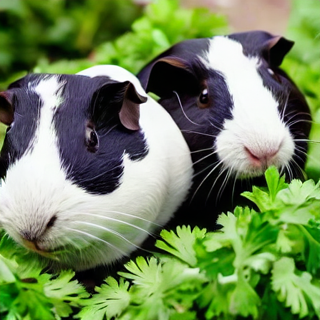} &
        \includegraphics[width=\imwidth]{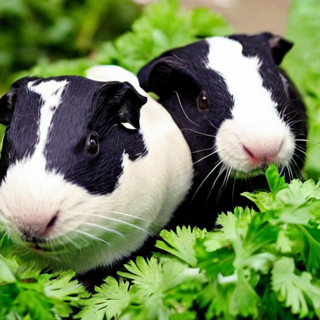} \\

         &
        \includegraphics[width=\imwidth]{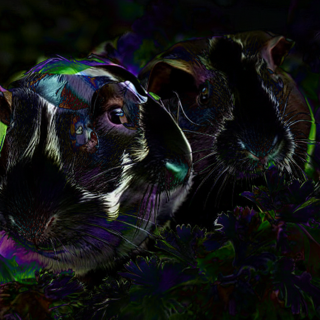} &
        \includegraphics[width=\imwidth]{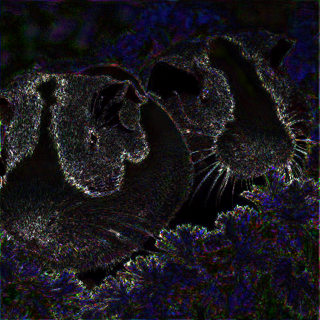} &
        \includegraphics[width=\imwidth]{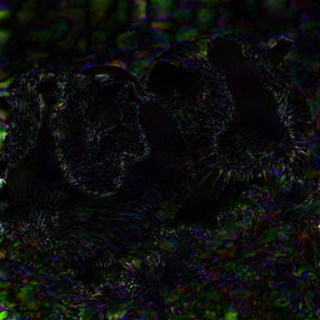} &
        \includegraphics[width=\imwidth]{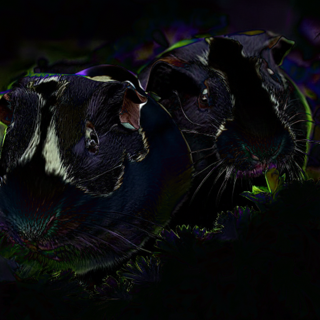} \\
        \midrule

        \multicolumn{5}{c}{\textit{Prompt: A shiba inu wearing a beret and \textcolor{red}{[black turtleneck $\mathcal{W}_*$]}}} \vspace{1mm} \\
        
        \includegraphics[width=\imwidth]{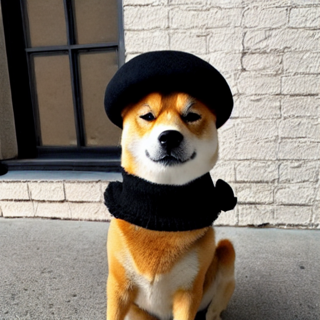} &
        \includegraphics[width=\imwidth]{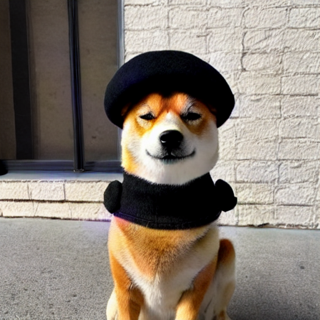} &
        \includegraphics[width=\imwidth]{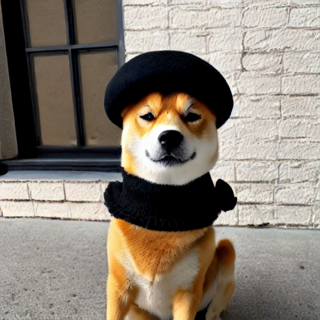} &
        \includegraphics[width=\imwidth]{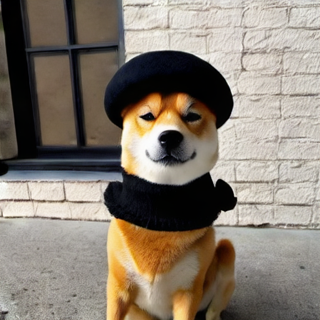} &
        
        \includegraphics[width=\imwidth]{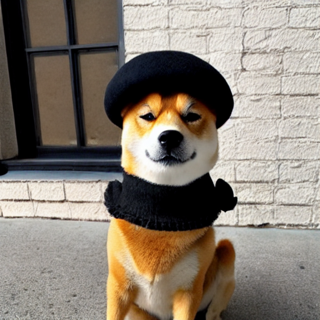} \\

         &
        \includegraphics[width=\imwidth]{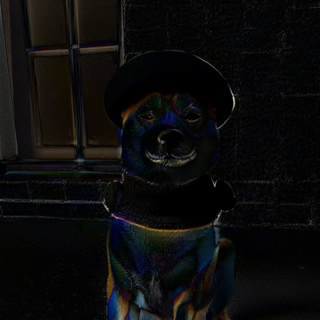} &
        \includegraphics[width=\imwidth]{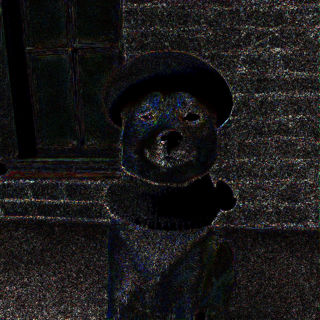} &
        \includegraphics[width=\imwidth]{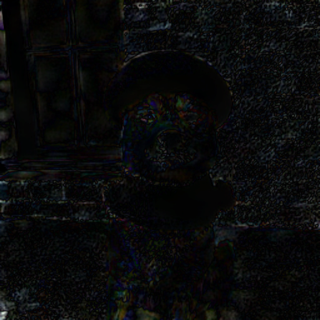} &
        \includegraphics[width=\imwidth]{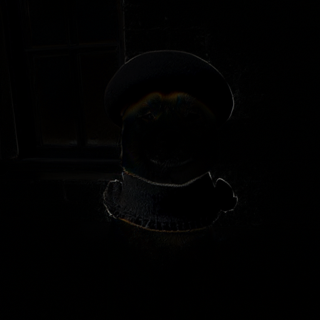} \\
        \midrule
        \end{tabular}
\caption{\label{fig:supp-watermark-part2} Qualitative results for different watermarking methods on generated images (part 2).}
\end{figure*}
\clearpage
\begin{figure*}[H]
\centering
    \scriptsize
    \newcommand{\imwidth}{0.165\textwidth}
        \setlength{\tabcolsep}{0pt}
        \begin{tabular}{c@{\hskip 2pt}cccc}
        \toprule
        Original & AquaLoRA & Stable Signature & RoSteALS & Ours (Object) \\
        \midrule

        \multicolumn{5}{c}{\textit{Prompt: A photo of a Ming Dynasty vase on a leather topped \textcolor{red}{[table $\mathcal{W}_*$]}.}} \vspace{1mm} \\
        
        \includegraphics[width=\imwidth]{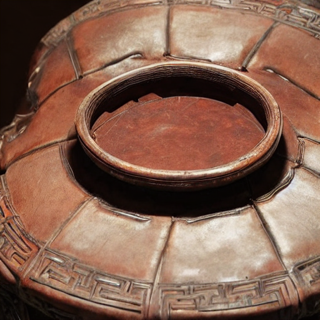} &
        \includegraphics[width=\imwidth]{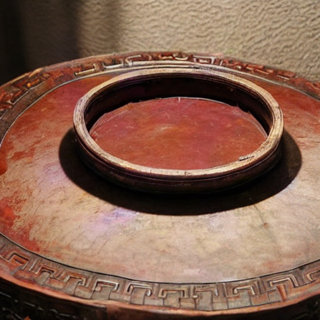} &
        \includegraphics[width=\imwidth]{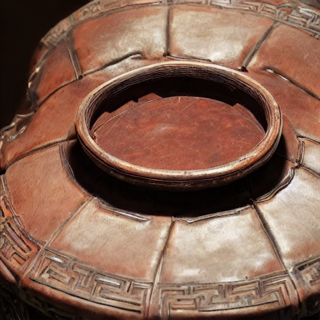} &
        \includegraphics[width=\imwidth]{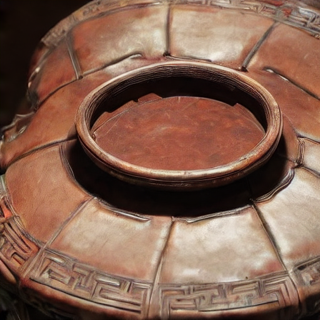} &
        \includegraphics[width=\imwidth]{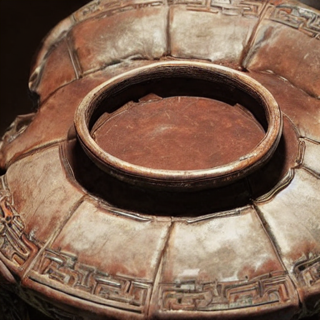} \\

         &
        \includegraphics[width=\imwidth]{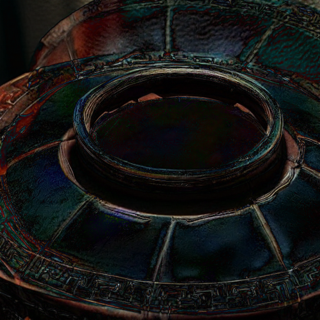} &
        \includegraphics[width=\imwidth]{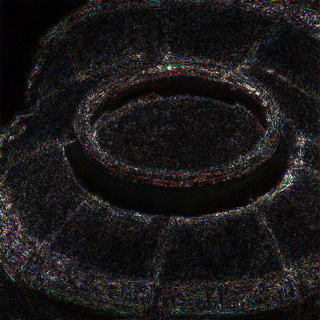} &
        \includegraphics[width=\imwidth]{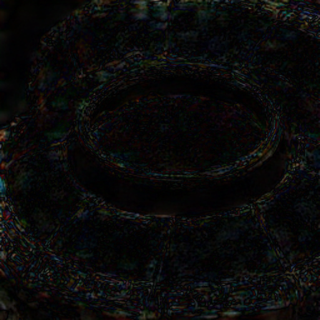} &
        \includegraphics[width=\imwidth]{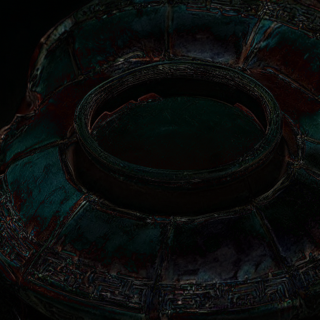} \\
        \midrule

        \multicolumn{5}{c}{\textit{Prompt: A snowy \textcolor{red}{[owl $\mathcal{W}_*$]} standing in a grassy field}} \vspace{1mm} \\

        \includegraphics[width=\imwidth]{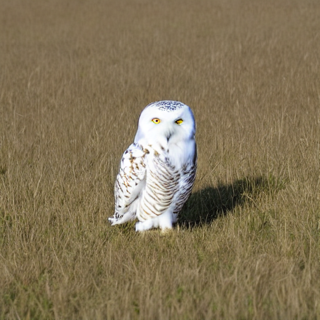} &
        \includegraphics[width=\imwidth]{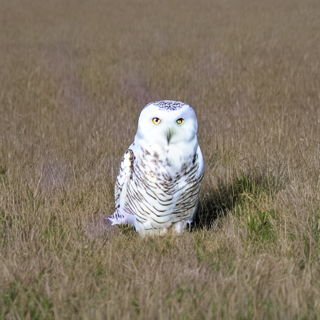} &
        \includegraphics[width=\imwidth]{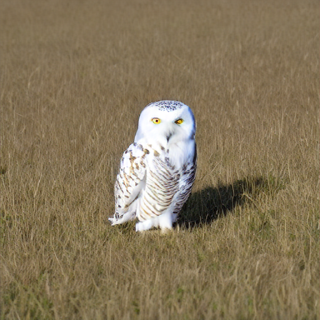} &
        \includegraphics[width=\imwidth]{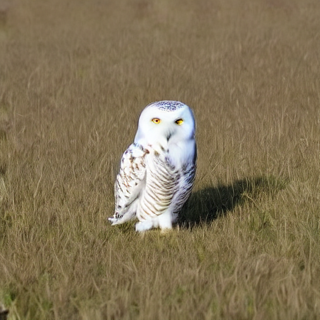} &
        \includegraphics[width=\imwidth]{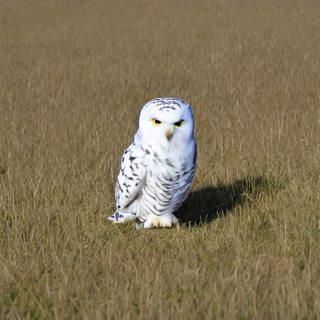} \\

         &
        \includegraphics[width=\imwidth]{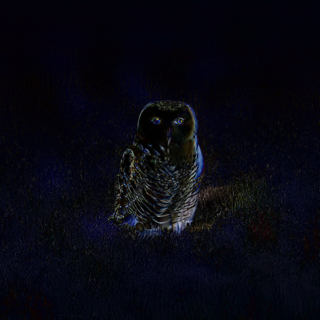} &
        \includegraphics[width=\imwidth]{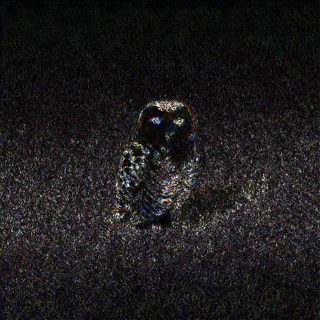} &
        \includegraphics[width=\imwidth]{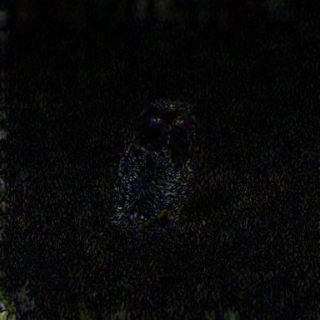} &
        \includegraphics[width=\imwidth]{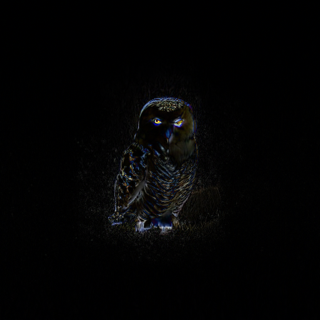} \\
        \midrule

        \multicolumn{5}{c}{\textit{Prompt: Background pattern with alternating roses and \textcolor{red}{[skulls $\mathcal{W}_*$]}}} \vspace{1mm} \\
        
        \includegraphics[width=\imwidth]{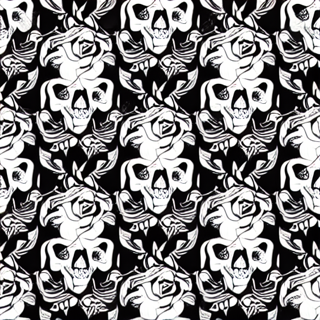} &
        \includegraphics[width=\imwidth]{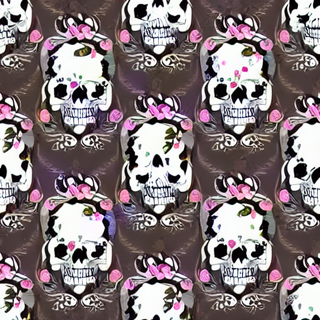} &
        \includegraphics[width=\imwidth]{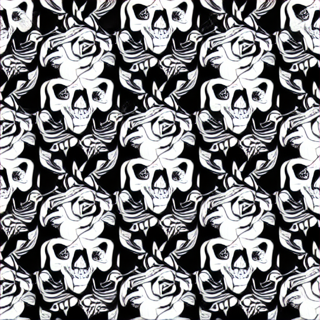} &
        \includegraphics[width=\imwidth]{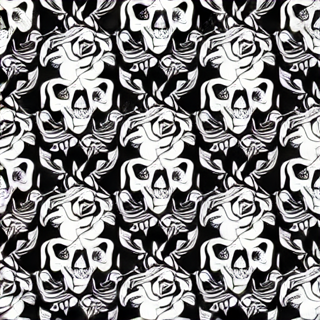} &
        \includegraphics[width=\imwidth]{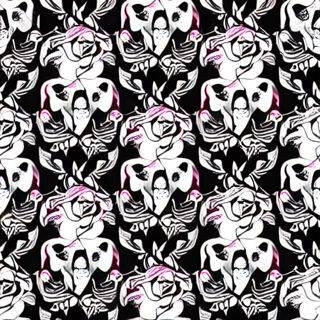} \\

         &
        \includegraphics[width=\imwidth]{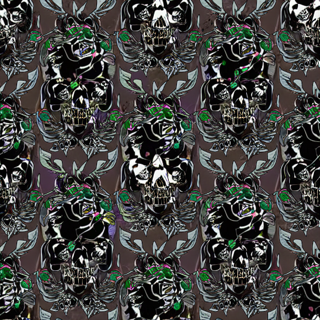} &
        \includegraphics[width=\imwidth]{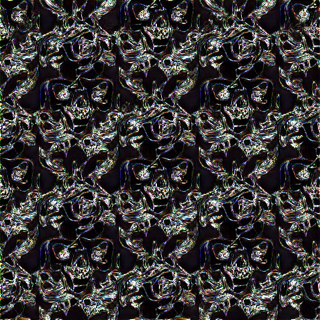} &
        \includegraphics[width=\imwidth]{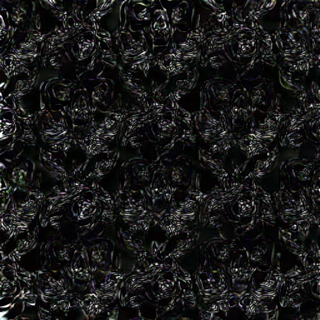} &
        \includegraphics[width=\imwidth]{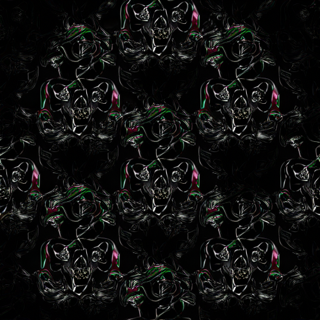} \\
        \midrule
        \end{tabular}
        \caption{\label{fig:supp-watermark-part2} Qualitative results for different watermarking methods on generated images (part 3).}
\end{figure*}
\clearpage
\begin{figure*}[!h]
\centering
    \scriptsize
    \newcommand{\imwidth}{0.165\textwidth}
        \setlength{\tabcolsep}{0pt}
        \begin{tabular}{c@{\hskip 2pt}cccc}
        \toprule
        Original & AquaLoRA & Stable Signature & RoSteALS & Ours (Object) \\
        \midrule

        \multicolumn{5}{c}{\textit{Prompt: A close-up of the keys of a \textcolor{red}{[piano $\mathcal{W}_*$]}}} \vspace{1mm} \\
        
        \includegraphics[width=\imwidth]{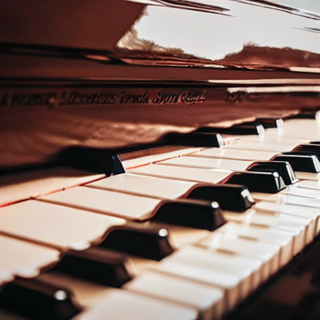} &
        \includegraphics[width=\imwidth]{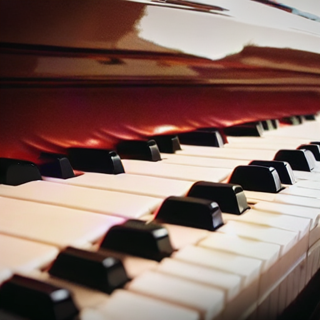} &
        \includegraphics[width=\imwidth]{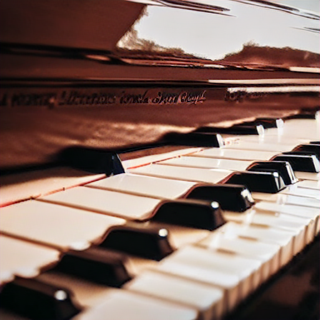} &
        \includegraphics[width=\imwidth]{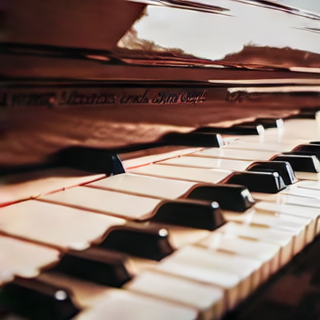} &
        \includegraphics[width=\imwidth]{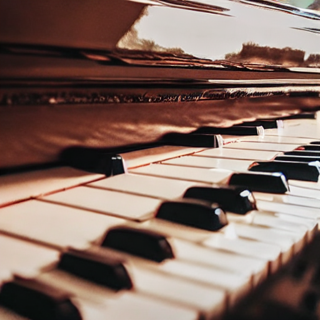} \\

         &
        \includegraphics[width=\imwidth]{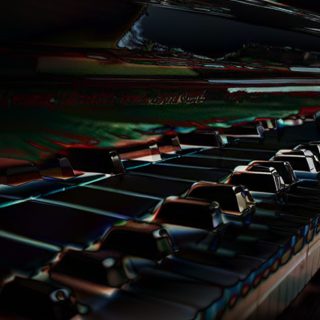} &
        \includegraphics[width=\imwidth]{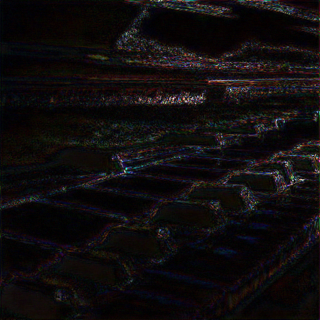} &
        \includegraphics[width=\imwidth]{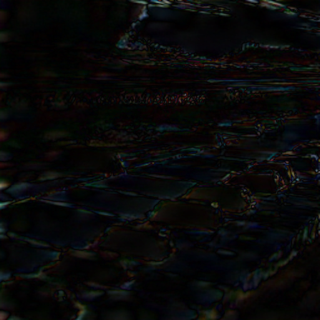} &
        \includegraphics[width=\imwidth]{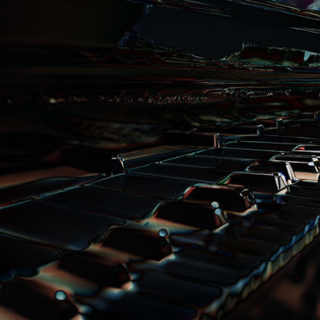} \\
        \midrule

        \multicolumn{5}{c}{\textit{Prompt: A map of \textcolor{red}{[Manhattan $\mathcal{W}_*$]}}} \vspace{1mm} \\
        
        \includegraphics[width=\imwidth]{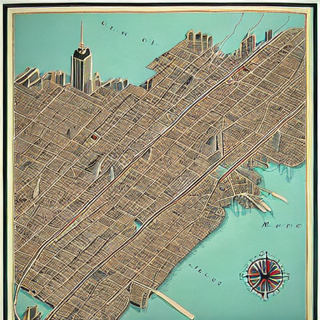} &
        \includegraphics[width=\imwidth]{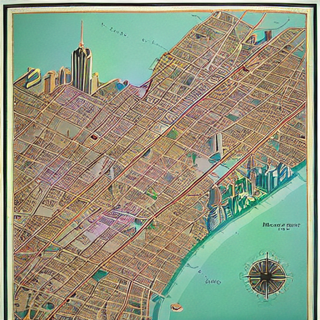} &
        \includegraphics[width=\imwidth]{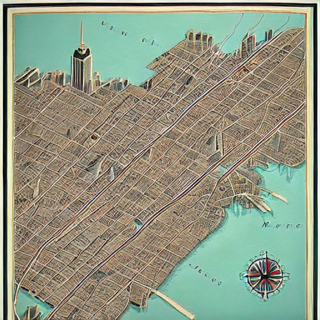} &
        \includegraphics[width=\imwidth]{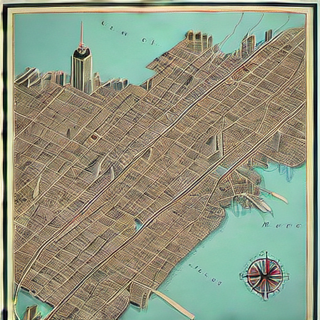} &
        \includegraphics[width=\imwidth]{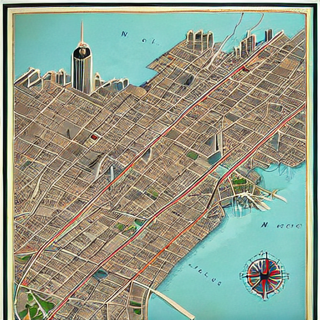} \\

         &
        \includegraphics[width=\imwidth]{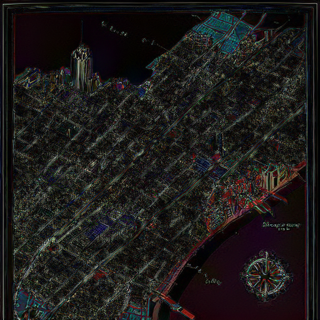} &
        \includegraphics[width=\imwidth]{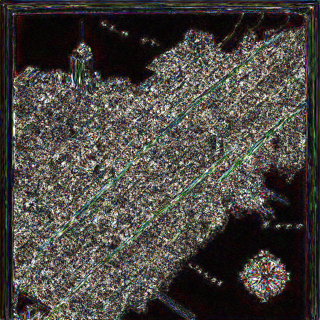} &
        \includegraphics[width=\imwidth]{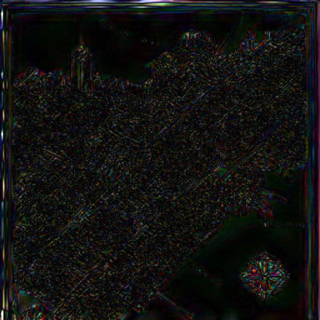} &
        \includegraphics[width=\imwidth]{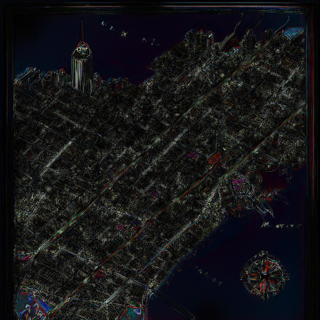} \\
        \midrule

        \multicolumn{5}{c}{\textit{Prompt: A \textcolor{red}{[sword $\mathcal{W}_*$]} in a stone}} \vspace{1mm} \\
        
        \includegraphics[width=\imwidth]{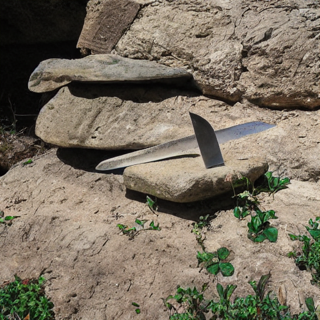} &
        \includegraphics[width=\imwidth]{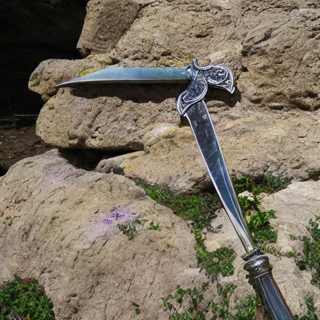} &
        \includegraphics[width=\imwidth]{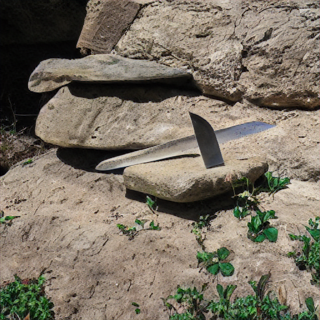} &
        \includegraphics[width=\imwidth]{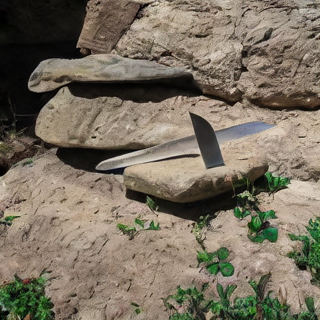} &
        \includegraphics[width=\imwidth]{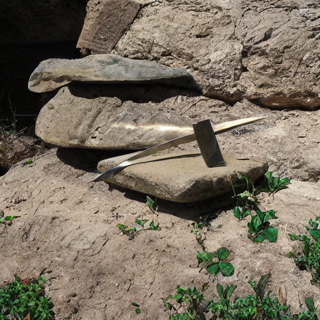} \\

         &
        \includegraphics[width=\imwidth]{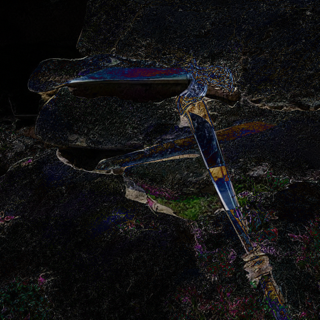} &
        \includegraphics[width=\imwidth]{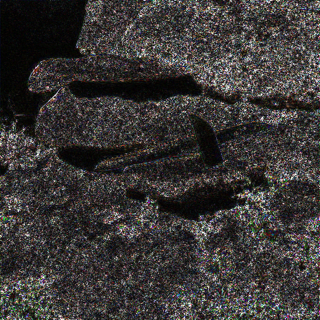} &
        \includegraphics[width=\imwidth]{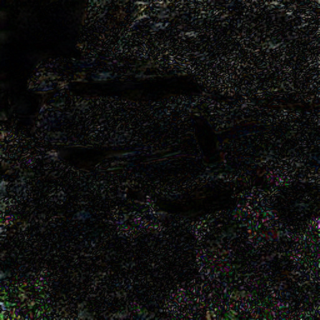} &
        \includegraphics[width=\imwidth]{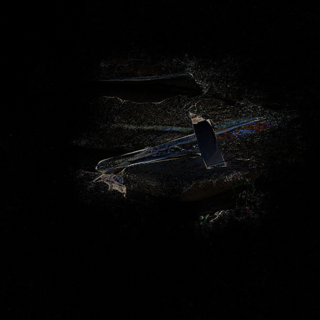} \\
        \midrule
        \end{tabular}
        \caption{\label{fig:supp-watermark-part2} Qualitative results for different watermarking methods on generated images (part 3).}
 
\end{figure*}

\clearpage

\begin{figure*}[!h]
\centering
    \scriptsize
    \newcommand{\imwidth}{0.165\textwidth}
    \setlength{\tabcolsep}{0pt}
    \begin{tabular}{c@{\hskip 2pt}cccc}
        \toprule
        Original & AquaLoRA & Stable Signature & RoSteALS & Ours (Whole Image) \\
        \midrule

        
        \includegraphics[width=\imwidth]{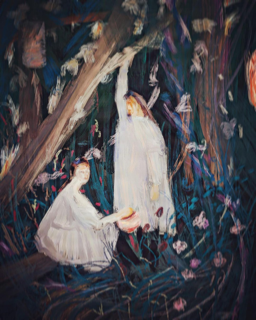} &
        \includegraphics[width=\imwidth]{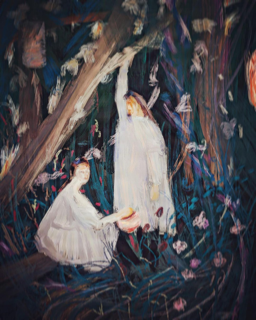} &
        \includegraphics[width=\imwidth]{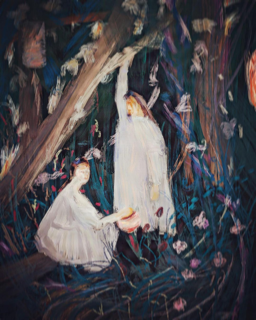} &
        \includegraphics[width=\imwidth]{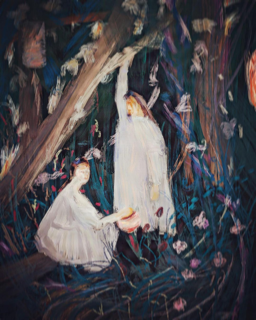} &
        \includegraphics[width=\imwidth]{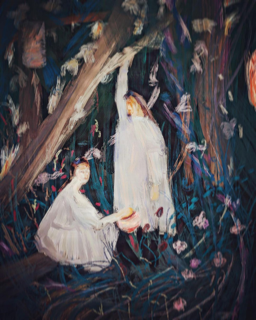} \\

         &
        \includegraphics[width=\imwidth]{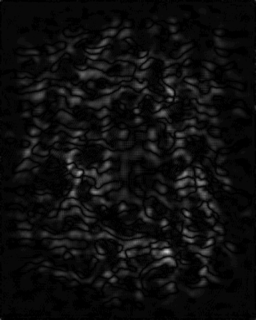} &
        \includegraphics[width=\imwidth]{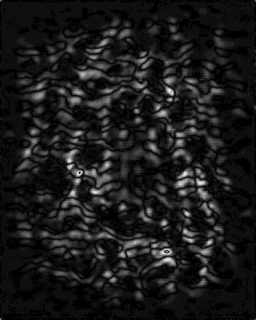} &
        \includegraphics[width=\imwidth]{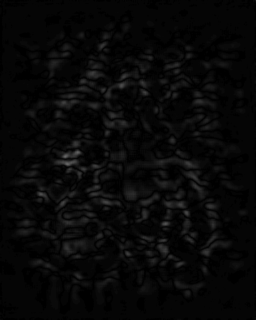} &
        \includegraphics[width=\imwidth]{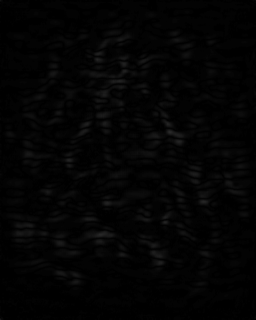} \\
        \midrule

        
        \includegraphics[width=\imwidth]{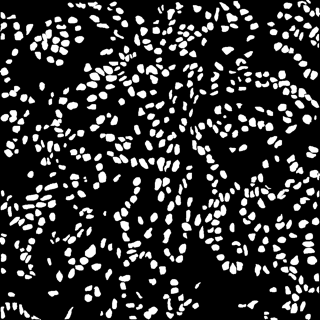} &
        \includegraphics[width=\imwidth]{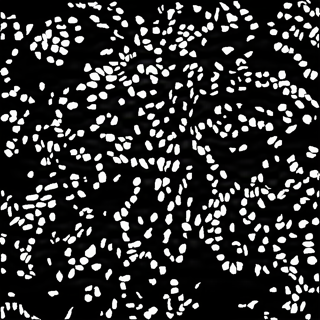} &
        \includegraphics[width=\imwidth]{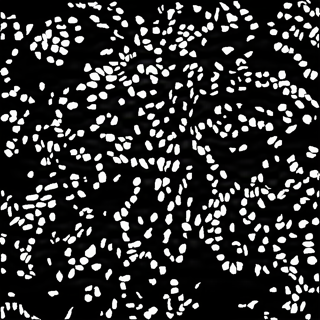} &
        \includegraphics[width=\imwidth]{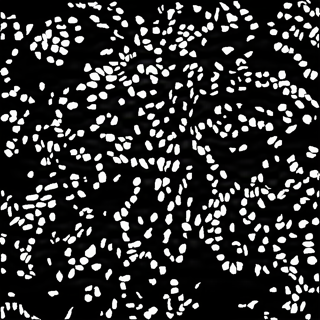} &
        \includegraphics[width=\imwidth]{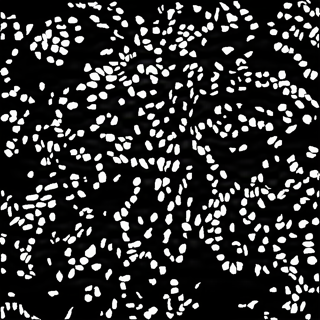} \\

         &
        \includegraphics[width=\imwidth]{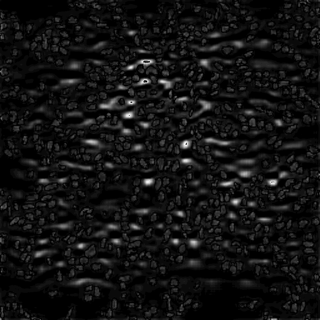} &
        \includegraphics[width=\imwidth]{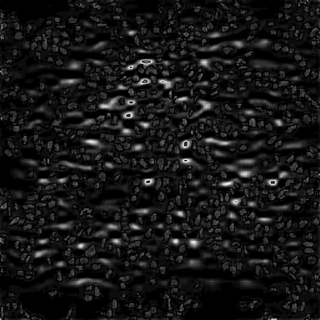} &
        \includegraphics[width=\imwidth]{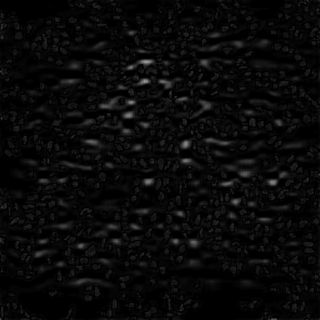} &
        \includegraphics[width=\imwidth]{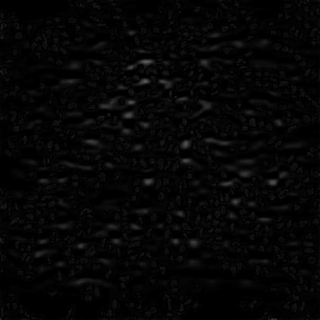} \\
        \midrule
    \end{tabular}
\caption{\label{fig:supp-watermark-part2} \textbf{Qualitative results for different watermarking methods on generated images} \emph{We present our results compared to the baselines on images from WikiArt \cite{artgan2018} and TNBC \cite{o2024survivalanalysisyoungtriplenegative} datasets.} }
\end{figure*}

\clearpage

\section{Attention Map of Watermarking Token} 
Our method appends a token $\mathcal{W}_*$ to the text prompt to integrate watermarking within the T2I generation. We utilize attention maps of each token in the text prompt to overlay the attention map of watermarking token $\mathcal{W}_*$ to control object-level watermarking. Here we provide the attention map of $\mathcal{W}_*$ token during T2I generation. We see that the difference image of watermarked and non-watermarked images is negligible. We amplify the difference image by $10$ and provide as the image in the last column.

\section{Choice of Latent loss for Trajectory Alignment}

Our method uses Latent loss $\|z_t(\mathcal{W}) - z_t^*\|^2$ (where $z_t(\mathcal{W})$ represent watermarked latents with $\mathcal{W}_*$ appended to the text prompt and $z_t^*$ denote non-watermarked latents with good image quality) for trajectory alignment during watermarking. In this section, we provide comparative of different loss functions to ensure trajectory alignment (preserve image quality) during watermarking while denoising. We see that in  \cref{fig:diffusion_pivotal_loss} loss maintains the best invisibility for watermarking. While diffusion loss performs better the case without any image loss, we observe that, for black-box watermarking, diffusion loss performs sub-par compared to the chosen latent loss that controls the trajectory of the latents.

\section{Robustness attacks implementation details}

We follow standard implementation for simulation of various attacks for watermarking as mentioned in \cite{10377226} and \cite{pmlr-v235-feng24k}. Crop 0.1 removes 10\% from the image retaining the center. Resize 0.2 scales the image to 20\% of its original size. Rotate 25 rotates the image by 25 degrees. Operations are performed using PIL and torchvision utilizing standard implementations from \cite{10377226, pmlr-v235-feng24k}.

\section{Medical image watermarking}

We tested our method to watermark medical images where invisibility of watermark is critical. Our method performs invisible watermarking our medical images with a high PSNR of \textbf{35.89} on the TNBC-Seg \cite{o2024survivalanalysisyoungtriplenegative} dataset while maintaining a bit accuracy of $\bf0.99$.



\section{Qualitative results compared to baselines}

In this section, we provide additional qualitative results compared to different watermarking methods that perform in-generation watermarking. We consider Stable Signature \cite{10377226}, AquaLoRA \cite{pmlr-v235-feng24k}, and RoSteALS \cite{bui2023rostealsrobuststeganographyusing} for comparing qualitative results.

\section{Heatmap detection based on user provided key}

We present the ability of our watermark heatmap generation module's ability to retrieve correct watermarked regions based on user key provided.

\begin{figure}[!t]
    \centering
    \includegraphics[width=0.8\linewidth]{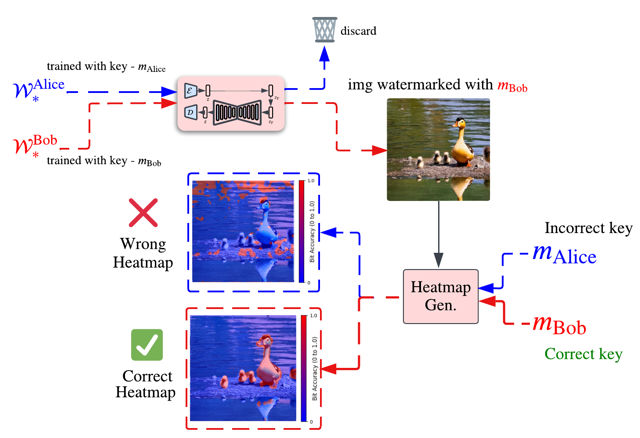}
    \caption{We show our watermark heatmap generation module's ability to retrieve heatmaps from watermarked images with different keys.}
    \label{fig:enter-label}
\end{figure}

